\pdfoutput=1
\documentclass[twoside]{article}

\usepackage[accepted]{aistats2021}

\usepackage[utf8]{inputenc} %
\usepackage[T1]{fontenc}    %
\usepackage{hyperref}       %
\usepackage{url}            %
\usepackage{booktabs}       %
\usepackage{amsfonts}       %
\usepackage{nicefrac}       %
\usepackage{microtype}      %
\usepackage{graphicx}
\usepackage{amsmath}
\usepackage{amsfonts}
\usepackage{multirow}

\newcommand{\methodName}{dGAP }
\newcommand{\LmethodName}{M-dGAP }
\newcommand{\YmethodName}{Y-dGAP }
\newcommand{\LYmethodName}{M-Y-dGAP }
\newcommand{\methodNameNS}{dGAP-NSS }
\newcommand{\methodNameNSparse}{dGAP-NSS w/o sparsity }
\newcommand{\methodNameGCN}{dGAP-GCN }

\newcommand{\LmethodNameNS}{M-dGAP-NSS }
\newcommand{\LmethodNameNSparse}{M-dGAP-NSS w/o sparsity }

\newcommand{\FullTitle}{Relate and Predict: Structure-Aware Prediction with Jointly Optimized Neural DAG}

\usepackage{amsmath}

\usepackage{graphicx}
\usepackage{xcolor}
\usepackage{algpseudocode}
\usepackage{float}
\usepackage{makecell}
\usepackage{amssymb}
\newcommand{\bX}{\boldsymbol{X}}
\newcommand{\bZ}{\boldsymbol{Z}}

\usepackage{caption}
\usepackage{bbm}
\usepackage[ruled,vlined]{algorithm2e}
\usepackage{etoolbox}

\newtoggle{appendix}
\togglefalse{appendix}

\newcommand{\fix}[1]{{{}}}

\usepackage{lipsum}
\usepackage{titlesec}

\titlespacing\section{0pt}{3pt plus 0pt minus 1pt}{2pt plus 1pt minus 1pt}
\titlespacing\subsection{0pt}{2pt plus 0pt minus 1pt}{0pt plus 0pt minus 0pt}
\titlespacing\subsubsection{0pt}{1pt plus 0pt minus 1pt}{0pt plus 0pt minus 0pt}
\titlespacing{\paragraph}{1pt}{1pt}{1pt}[0pt]  
\usepackage{etoolbox}
\makeatletter
\preto{\@tabular}{\parskip=3pt}
\makeatother

\allowdisplaybreaks

\usepackage{enumitem}
\setlist[itemize]{leftmargin=*}

\begin{document}
\graphicspath{{figures/}}

\twocolumn[

\aistatstitle{\FullTitle}
\aistatsauthor{Arshdeep Sekhon, Zhe Wang \& Yanjun Qi}  
\aistatsaddress{Department of Computer Science\\
University of Virginia\\
Charlottesville, VA 22903, USA \\
\texttt{\{as5cu,zw6sg,yanjun\}@virginia.edu}  \\
}

]

\begin{abstract}
 
Understanding relationships between feature variables is one important way humans use to make decisions. However, state-of-the-art deep learning studies either focus on task-agnostic statistical dependency learning or do not model explicit feature dependencies during prediction.
We propose a deep neural network framework, \methodName, to learn neural \underline{d}ependency \underline{G}raph and optimize structure-\underline{A}ware target \underline{P}rediction simultaneously. 
\methodName trains towards a structure self-supervision loss and a target prediction loss jointly. Our method leads to an interpretable model that can disentangle sparse feature relationships, informing the user how relevant dependencies impact the target task.  
We empirically evaluate  \methodName on multiple simulated and real datasets.  \methodName is not only more accurate, but can also recover correct dependency structure.

\end{abstract}

\section{Introduction}

Cognitive psychologists have identified relational structure as one primary component humans rely on to tackle unstructured problems \cite{halford2010relational}. Relational representations are the foundation in higher cognition. One primary relational thinking describes complex systems as compositions of entities and their interaction graphs. In this paper, we borrow such an idea and design graph-oriented relational representation learning into state-of-the-art deep neural networks.  Learning such structure representations from data can provide semantic clarity, ease of reasoning for generating new knowledge,  and possibly causal interpretation.

Existing deep learning literature has proposed effective ways, like using graph neural networks, to represent data when relational graphs are known apriori~\cite{zhou2018graph}. However, little attention has been paid to address cases when the underlying relation graph is unknown. We, therefore, ask a question: is it possible to learn graph-based relational knowledge from data, for both knowledge communication and prediction using knowledge at the same time.

Our method, we name \methodName, jointly learns neural \underline{d}ependency \underline{G}raph and exploits the inferred graph for structure-\underline{A}ware target \underline{P}rediction. \methodName is trained end-to-end by optimizing the task-specific loss plus a structure self-supervision loss as regularization.  Our method extracts knowledge of a target task at a macro level, and at the same time using the learned knowledge to conduct reasoning. In summary, we make the following contributions: 
\begin{itemize}
    \item \methodName exploits an explicit neural dependency module to model and learn variable interactions relevant to a task we care. 
    Our design is in a direction towards making deep learning more human-like, since our neural dependency network-oriented design is consistent with the way human organize knowledge for higher cognition. 

    \item Our method leads to an interpretable model that can disentangle sparse feature relationships, informing the user how relevant dependencies impact the target task. Having a structure not only helps understand the problem at a macro level, it also helps to pinpoint areas that require deeper understanding. 

    \item We empirically evaluate  \methodName on multiple simulated and real-world datasets.  \methodName predicts accurately and can discover task-oriented knowledge. On simulated cases we empirically prove that the discovered graphs match well with the true dependency networks, outperforming state-of-the-art baselines with a significant gain.
\end{itemize}
Like a tourist needs a map, human thinking requires structure. We borrow such an intuition in the design of \methodName. To the authors' best knowledge, \methodName is the first deep learning architecture that extracts knowledge in the form of neural dependency graph and conducts knowledge based predictions at the same time.

Using simulated and real-world datasets, we show that \methodName is accurate, achieving higher prediction performance compared to baselines that do not account for feature relationships. Further, we show that \methodName can provide interpretable knowledge of which features interact relevant for prediction. Owing to \methodName's ability to select interactions relevant for prediction, we show that \methodName is data efficient as well as more robust to noisy features.

\section{Background: Dependency Structure Learning}

Learning the dependency structure between input variables is an important task in machine learning. %
Traditionally, statistical Graphical Models intuitively represent relationships between random variables as conditional dependence graphs. In detail, the variables of interest are represented by a set of nodes $V$, with relationships as edge set $E$ in graph $\mathcal{G}=(V,E)$. The edges $E$ represent conditional dependencies between the variables. Essentially, learning these relationships is equivalent to learning a factorization of the joint probability distribution. %
One notable method  that uses representation of undirected dependency graphs for a joint probability distribution is the Markov Random Field(MRF).  %

An MRF models a set of random variables to have a joint probability stipulated by an undirected graph $\mathbb{G}=(\mathbb{V}, \mathbb{E})$. These models satisfy the markov properties: (1) Pairwise Markov Property: Any two non-adjacent variables are conditionally independent given all other variables:  (2) Local Markov Property : A variable is conditionally independent of all other variables given its neighbors. (3) Any two subsets of variables are conditionally independent given a separating subset. 
Discovering the MRF dependency structure $\mathcal{G}$ is a reconstruction of the graphical structure of a Markov Random Field  from independent and identically distributed samples.

Assuming a dataset $D=(\boldsymbol{x}^1, \dots, \boldsymbol{x}^n)$, whose $\boldsymbol{x}^s\ \in \mathbb{R}^{p}$, that is sampled i.i.d. from the parametric distribution $p_{\theta}$, a maximum likelihood estimator is defined as $\boldsymbol{\theta}_n$: \begin{equation}
    \ell_n(\boldsymbol{\theta}_n;D)=\sum_{s=1}^{n}\log p_{\boldsymbol{\theta}}(\boldsymbol{x}^s)
\end{equation}

To make the learning process more tractable, \cite{besag1977efficiency} proposed to use pseudo likelihood as an approximation. Pseudo-log likelihood\cite{besag1977efficiency} offers a tractable replacement for the likelihood, defined as $\ell_p$:
\begin{equation}
    \ell_p(\boldsymbol{\theta}_p;D)=\sum_{s=1}^{n}\sum_{i=1}^{p}\log p_{\boldsymbol{\theta}}(\boldsymbol{x}^s_i|\boldsymbol{x}^s_{-i} )
\end{equation}
In case of a Markov Random Field, this reduces to:
\begin{equation}
    \ell_p(\boldsymbol{\theta}_p;D)=\sum_{s=1}^{n}\sum_{i=1}^{p}\log p_{\boldsymbol{\theta}}(\boldsymbol{x}^s_i|\boldsymbol{x}^s_{N(i)} )
    \label{eq:pseudo_mrf}
\end{equation}
Here $N(i)$ stands for neighbors of node $i$ in the dependency structure $\mathbb{G}$.

\section{Method}
{\bf Notations:}  We denote the input data matrix by $\boldsymbol{X}\in \mathbb{R}^{n \times p}$ where $n$ represents the number of samples, and $p$ represents the number of input nodes or variables. We represent the corresponding labels $\boldsymbol{Y}\in \mathbb{R}^{n\times C}$, where $C$ indicates the number of classes. We denote the discrete binary graph between the variables by $\boldsymbol{Z}\in \mathbb{R}^{p \times p}$. We denote one individual sample and its corresponding label by $\boldsymbol{x}, \boldsymbol{y}$.

 Our main objective is to learn undirected dependencies between input variables jointly with  a target supervised task. Accordingly, \methodName has two modules: a Structure Learner $\mathbb{S}$ and a Task Learner $\mathbb{T}$. In the following subsections, we describe in detail these two modules.  %
\begin{figure*}
    \centering
    \includegraphics[width=\textwidth]{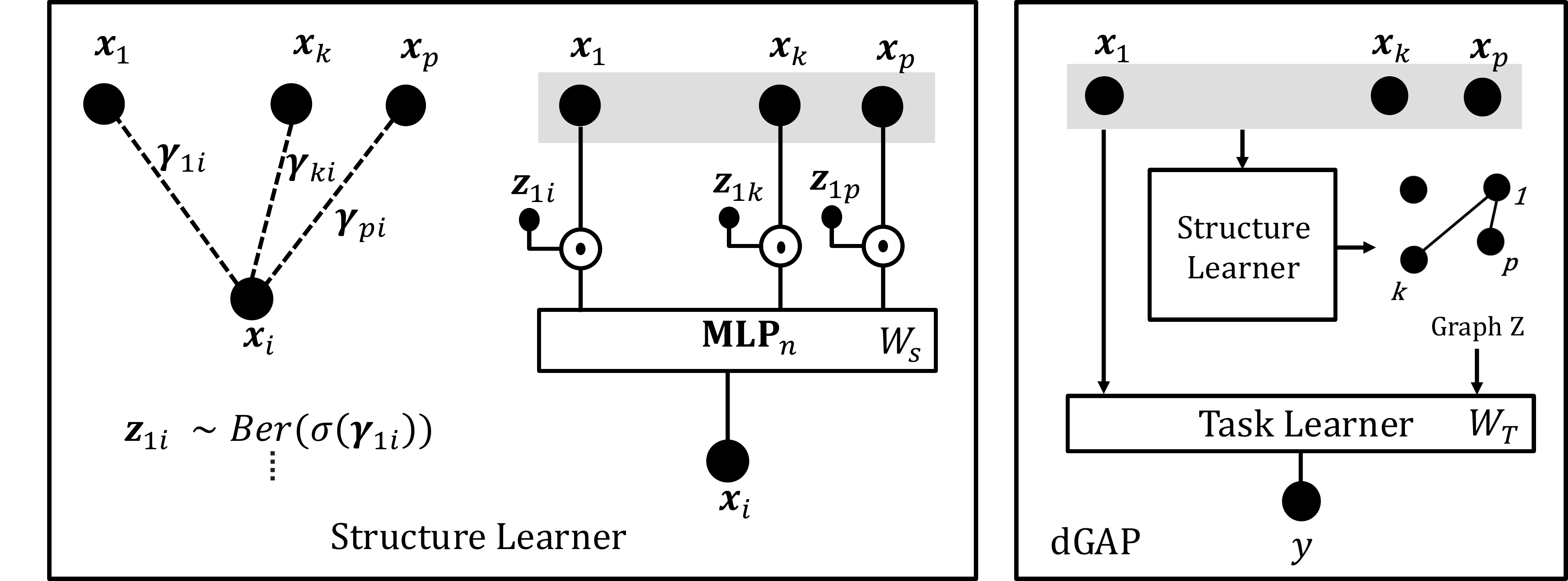}
    \caption{A schematic diagram representing \methodName. }
    \label{fig:model_full}
\end{figure*}

\subsection{Structure Learner ($\mathbb{S}$)}
The goal of this module is to learn graph-oriented structure representation about the input variables. We choose to learn generic undirected dependency graphs  via a neural network module.  

To learn these structures, we introduce binary variables $\boldsymbol{Z}=\{\boldsymbol{Z}_{i,j}\}$, that represent the dependency structure between the relevant variables (aka a neural dependency graph $\bZ$). $\boldsymbol{Z}_{i,j}$ specifies the presence or absence of a dependency between variable $i$ and $j$. Thus, the state of without a dependency can be encoded by multiplying the corresponding parent node by 0. The neighbors of each input variable $i$ is given by $\boldsymbol{Z}_{:,i}$. Finding the optimal dependency structure corresponds to a search over all configurations of $\boldsymbol{Z}$.  Besides we need to optimize the other learnable model parameters for each possible dependency structure in the search space. 
To avoid this intractable procedure, we parametrize $\boldsymbol{Z}$ by placing a distribution over $\boldsymbol{Z}$. This enables us to directly optimize the parameters
of this distribution to arrive at a data-driven optimal dependency structure. In detail, $p_{\boldsymbol{\theta}}(\boldsymbol{x},\boldsymbol{Z})=p_{\boldsymbol{\theta}}(\boldsymbol{x}|\boldsymbol{Z})p(\boldsymbol{Z})\delta_{\boldsymbol{Z},\boldsymbol{Z}'}$ where $\delta_{\boldsymbol{Z},\boldsymbol{Z}'}$ is the Kronecker delta, which effectively selects a single structure. Connecting from Eq.~\ref{eq:pseudo_mrf}, to optimize the pseudo log likelihood of the data distribution using dependency structure $\boldsymbol{Z}$:
\begin{equation}
    \ell_{p}(\boldsymbol{\theta})=\sum_{s=1}^n\log (p_{\boldsymbol{\theta}}(\boldsymbol{x}_s|\boldsymbol{Z}))\approx \sum_{s=1}^n\sum_{i=1}^{p} \log p_{\boldsymbol{\theta}}(\boldsymbol{x}_{si}|\boldsymbol{x}_{s,-i}, \boldsymbol{Z}_{:,i})
    \label{eq:joint_graph_prob}
\end{equation}
Now, we treat
each $\boldsymbol{Z}_{i,j}$ as an independent random variable, sampled from a Bernoulli distribution with mean $\gamma_{i,j}$,
i.e. $\boldsymbol{Z}_{i,j} \sim p(\boldsymbol{Z}_{i,j} )$ = Ber($\boldsymbol{Z}_{i,j}$ ). We denote the mean parameter of this Bernoulli distribution for all edges as a matrix $\boldsymbol{\gamma}$. 
In detail, we introduce a learnable parameter matrix $\boldsymbol{\gamma}\in \mathbb{R}^{p\times p}$.  The sigmoid of the $ij^{th}$ entry in this matrix, $\sigma(\boldsymbol{\gamma}_{ij})$,  indicates the Bernoulli probability of an interaction edge between the input $i^{th}$ and $j^{th}$ node. 
 We use $\boldsymbol{Z}\sim(Ber(\sigma(\boldsymbol{\gamma})))$ defines binary graphs, representing dependencies between the input features.

In summary, the graph $\boldsymbol{Z}$ acts as a mask to reconstruct the input in a self supervision setting. We reconstruct $\boldsymbol{x}_i$ from $[\boldsymbol{Z}_{j,i}\odot x_{j}] \{j = \{1, \dots,p\},\  j \neq i\}$. In other words,  for feature $i$, we take all other features as input features masked by $\boldsymbol{Z}_{:,i}$ to predict feature $i$ via a function (MLP in our case).  Mathematically, $\boldsymbol{x}^1_i =[ \boldsymbol{Z}_{1,i}\boldsymbol{x}_{1}, \cdots, \boldsymbol{Z}_{p,i}\boldsymbol{x}_{p} ], \forall i \in \{1, \dots, p\}, i\neq j$.

\paragraph{Self-supervision loss:} We use $\boldsymbol{x}^1_i$, obtained using the input masked by the neighbors learnt by dependency graph  $\boldsymbol{Z}_{:,i}$, as input to an MLP to reconstruct $\boldsymbol{x}_i^{'}$: $\boldsymbol{x}_i=MLP_i(\boldsymbol{x}^1_i)$. We use reconstruction loss $\ell_{struct}=\sum_{i=1}^p ReconstructionLoss(\boldsymbol{x}_i, \boldsymbol{x}^{'}_i)$. If $\boldsymbol{x}_i\in \mathbb{R}$, we use Mean Squared Error Loss for this self-supervision based reconstruction. This assumes Gaussian distribution for $p_{\boldsymbol{\theta}}(\boldsymbol{x}_{i}|\boldsymbol{x}_{-i}, \boldsymbol{Z}_{:,i})$ in Eq~\ref{eq:joint_graph_prob}. When on categorical input variables $\boldsymbol{X}_i\in \mathbb{R}^{L}$, where $L$ is the number of categories in the categorical variable $\boldsymbol{X}_i$, we use Negative Log Likelihood Loss as Reconstruction loss. Essentially we assume multinoulli distribution for $p_{\boldsymbol{\theta}}(\boldsymbol{x}_{i}|\boldsymbol{x}_{-i}, \boldsymbol{Z}_{:,i})$ in Eq~\ref{eq:joint_graph_prob}.

\paragraph{M-{\methodName}:} $\boldsymbol{Z}$ may differ depending on data samples. While some interactions may be present in all the samples irrespective of the target class it belongs to, some interactions may be specific to a cluster of samples.  To incorporate one kind of this variability, we use the training data labels to construct class specific graphs. For example, in the case of classification, if $\boldsymbol{y}\in\{1,\dots, C\}$, we introduce a total of $C$ learnt dependency graphs,  $\boldsymbol{Z}_c$ where $c \in \{1,\dots, C\}$, where $C$ is the number of classes.  During the training step,  we use the sum of the class specific losses: $\ell_{struct}=\sum_{c=1}^{C}\mathbbm{1}_{\boldsymbol{y}=c}\ell_{struct}(\boldsymbol{x}, \mathbb{S}_c(\boldsymbol{x}))$  We denote this variation as MultiGraph-\methodName (short as M-\methodName). 

\subsection{Task Learner ($\mathbb{T}$)}

We want to explicitly incorporate the dependency structures  into our target objective. We want to encourage discovery of interactions that are specifically relevant to the end task.  %
For this purpose, we use Message Passing Neural Networks (MPNN) as an interface to jointly represent dependency structures and input features.

For each specific sample, 
in the first step, we represent each of its $i$-th feature node $\boldsymbol{x}_i$ as a position-specific embedding vector $\boldsymbol{h}^0_i$: $\boldsymbol{h}^0_i=[\boldsymbol{x}_i||\boldsymbol{W}^{pos}_i|| \boldsymbol{x}]$, where $\boldsymbol{W}^{pos}\in \mathbb{R}^{{p} \times d_{pos}}$.  Here, $[\odot || \odot ]$ denotes concatenation.
 After these node specific embeddings,  we use graph attention to combine the discrete graph $\boldsymbol{Z}$ and input embeddings $\boldsymbol{h}^0_i\ i \in \{1,\dots,p\}$. In detail, we learn a graph attention\cite{velivckovic2017graph} using the following equation for each edge  $\boldsymbol{Z}_{i,j}$:
    \begin{equation}
    \boldsymbol{\alpha}_{ij,k}^{l}=\dfrac{exp(LeakyReLU(\boldsymbol{a_e^T}[\boldsymbol{W}^l_k\boldsymbol{h}^{l-1}_i|| \boldsymbol{W}_k^l\boldsymbol{h}^{l-1}_j]))}{\sum_{j=1}^p exp(LeakyReLU(\boldsymbol{a_e^T}[\boldsymbol{W}_k^l\boldsymbol{h}^{l-1}_i|| \boldsymbol{W}_k^l\boldsymbol{h}^{l-1}_j]))}
        \label{eq:attn}
    \end{equation}
    \begin{equation}
        \boldsymbol{h}_i^{l}=\sigma(\sum_{j\in N_i} \boldsymbol{\alpha}_{ij,k}^{l}\boldsymbol{W}_k^l \boldsymbol{h_j}^{l-1})
    \end{equation}
Via the above equation, the embedding of node $i$ is updated by an attention weighted sum of its neighbors' embedding. More specifically, we use a multi head attention based formulation:
    \begin{equation}
        \boldsymbol{h}_i^{l}= ||_{k=1}^{K}\sigma(\sum_{j \in N_i} \boldsymbol{\alpha}_{ij,k}^l\boldsymbol{W}^l_k \boldsymbol{h}_j^{l-1})
        \label{eq:embed}
    \end{equation}
    Here, the neighbors for a node $i$ are obtained using $\boldsymbol{Z}_{:,i}$, $k$ denotes the $k$-th attention head out of a total of $K$ heads, and $||$ denotes concatenation including all $K$ heads based embedding. $l$ represents the index of graph attention layers (steps).  Eq. ~\ref{eq:attn} and Eq. ~\ref{eq:embed} are repeated for $L$ layers. This gives us the final representation $\boldsymbol{h}_{L}\in \boldsymbol{R}^{p \times d}$.
  To represent all the nodes together for a sample $\boldsymbol{x}$, we use a CLS node based pooling\cite{devlin2018bert}. That is, we add an additional node that is connected to all other nodes in the graph. The final representation of this node, followed by a single layer MLP and a log softmax,  makes the final graph level prediction $\hat{\boldsymbol{y}}$. 
    The loss from the end task $\ell_{task}=\ell_{pred}({\boldsymbol{y}},\hat{\boldsymbol{y}})$.

     \paragraph{M-{\methodName}:}For the case of multiple graphs variation,we use multiple Graph Attention Networks one for each of $\boldsymbol{Z}_c\forall c \in \{1, \dots, C\}$.  This gives us output representations $\boldsymbol{h}^{L}_{{CLS}_c}$, one for each class. For the final classification, we add another attention layer on the $c$ representations. In detail, we use a context vector $\boldsymbol{v}\in \mathbb{R}^{d}$. We obtain attention scores $\beta_c=Softmax_{1,\dots,c}(\boldsymbol{v}\odot \boldsymbol{h}^{L}_{{CLS}_c})$. Finally, we use $\boldsymbol{h}_{out}=\sum_{c=1}^C \beta_c\boldsymbol{h}^{L}_{{CLS}_c}$ as input to the MLP layer for classification.  
    \subsection{Training and Loss} We additionally use a sparsity constraint $\ell_{sparse}=\sum_{i=1}^p \sum_{j=1}^p \sigma(\gamma_{ij})$ as a regularization to  encourage learning of a sparse $\boldsymbol{Z}$. 
   To sample discrete graphs, we use a gumbel softmax \cite{jang2016categorical, maddison2016concrete} trick to sample from Bernoulli distribution parametrized by $\sigma(\boldsymbol{\gamma})$. 
   \begin{equation}
   \small
        \boldsymbol{Z}_{i,j} = \dfrac{exp(\log(\sigma(\boldsymbol{\gamma}_{ij})+\epsilon_1)/\tau)}{exp(\log(\sigma(\boldsymbol{\gamma}_{ij})+\epsilon_1)/\tau+exp(\log(1-\sigma(\boldsymbol{\gamma}_{ij})+\epsilon_2)/\tau}
    \end{equation}
    where $\epsilon_1$ and $\epsilon_2$ are i.i.d samples drawn from a Gumbel(0, 1) distribution and $\tau$ is a temperature
parameter. The Gumbel-Softmax distribution is differentiable for $\tau>0$. This allows us to sample discrete graphs for the graph attention network, while being able to propagate gradients to the learnt parameter $\boldsymbol{\gamma}$ both from the structure learner ($\boldsymbol{W}_S$) as well as the task learner parameters ($\boldsymbol{W}_T$). 
 To optimize the model, we minimize $\ell$  using Adam optimizer{} to train all these components: graph parameter $\boldsymbol{\gamma}$, structure learner ($\boldsymbol{W}_S$) as well as the task learner parameters ($\boldsymbol{W}_T$) together.
 
   \begin{equation}\ell = \ell_{task} + \lambda_{struct}\ell_{struct} + \lambda_{sparse}\ell_{sparse}    
   \end{equation}

For cases where an undirected graph does not fit the data, we add an additional constraint $\lambda_{dag} = \sum_{i\neq j}=cosh(\sigma(\gamma_{ij} )\sigma(\gamma_{ji}))$, that suppresses length-2 cycles\cite{ke2019learning}. 

We note that using a binary graph instead of the direct dense $\sigma(\boldsymbol{\gamma})$ as the dependency graph allows us to have a probabilistic interpretation of the graphs. Additionally, the combination of a global adjacency graph along with sample level attention allows us to learn global binary dependency  graphs with local (sample level) fine-tuning regarding which edges were most important for prediction.

\subsection{Connecting to related studies}

One classic statistical way to learn dependency relationships from data was from the family of probabilistic graphical models (PGM). These models represent the random variables as a node set $V$ and the relationships between variables as an edge set $E$. Edges denote conditional dependencies among variables. Learning PGM is equivalent to learning a factorization of the joint probability distribution. PGM roughly fall to two groups: (1) undirected PGM, for which a non exhaustive list includes Gaussian Graphical Models, Ising Models, etc; (2) Directed PGM that tries to model and learn directed conditional dependency structure, normally plus acyclic graph constraints. A few recent studies \cite{lachapelle2019gradient,zheng2018dags,zheng2019learning}  learned directed dependencies via deep neural networks (DNN). In many cases of PGM learning, sparsity regularization is used to encourage a final sparse graph. Besides, few recent literature aims to learn causal Bayesian networks~\cite{magliacane2018domain,ke2019learning} via DNN based formulations. In the absence of intervention distributions, these causal methods can only learn a Markov equivalence class of the true networks. All studies in this category perform unsupervised estimation, that is they don't jointly optimize a task-specific loss when learning the dependency structures from data.

A few recent DNN studies also aimed to estimate relationships among input variables, but their relations are not about dependency structure. For instance, Neural Relational Inference (NRI)~\cite{kipf2018neural}, is a neural network based interaction learning framework designed for modeling dynamical interacting systems. NRI learns sample-specific pairwise interactions between input units using a physics perspective. NRI represented the edges as latent variables in a Variational Autoencoder(VAE) framework. Relational graphs in NRI are sample-conditioned and are learned via unsupervised alone, that is  the learnt interactions are not informed by a target task objective. Differently, another recent study, LDS from \cite{franceschi2019learning} proposed to jointly learn the parameters of graph convolutional networks (GCNs) and  the graph structure (as unknown hyperparameter of GCN) by approximately solving a bilevel program. The whole learning is guided by a downstream node classification objective. Comparing to ours, the graph LDS  has no dependency semantics and the whole formulation only applies to GCN.

Our method also connects to an array of methods that tried to disentangle variable interactions from deep neural models from a post-hoc interpretation perspective. For instance, \cite{tsang2017detecting} detects interactions learnt by multilayer perceptron networks (MLP) by decomposing the weights of the MLP. Similarly, \cite{cui2019recovering} estimates global pairwise interaction effects of a MLP model using Bayesian parameter analysis. Another closely related work \cite{tsang2017detecting, tsang2020feature} focused on discovering interactions between sparse features and then explicitly encoding them for explaining MLP based recommendation neural models in a two stage approach. All of these methods are post-hoc interpretation methods, and most are restricted to specific DNN architectures. Loosely related, another set of feature interaction based interpretation studies have been proposed in the literature to explain decision-tree based learning methods  \cite{friedman2008predictive,sorokina2008detecting,lundberg2018consistent}. Notably, two recent studies from the post-hoc interpretation  category stack the discovered interactions~\cite{tsang2017detecting,tsang2020feature} into a second stage of prediction task in order to improve prediction performance.  These methods are different from \methodName, because our method trains end-to-end to discover and utilize the  interactions for the task at hand jointly.

Besides, most of the post-hoc interaction-based interpretation methods focused on learning feature interactions for each sample , like \cite{janizek2020explaining}. \methodName, instead, provides global dependency explanations. Global explanations are desirable for interpretable data and model summarization\cite{tsang2017detecting}. %

\section{Experiments}

\paragraph{Data: } We group our experiments regarding on simulation and real world datasets. The prediction tasks include both regression and classification. On 3 simulated and 7 real-world datasets, we show {\methodName} can achieve state-of-the-art prediction performance, and also discover meaningful dependencies in the data. 
The real-world datasets covered our experiments include both tabular inputs and image inputs.

\paragraph{Evaluation Metrics:} When evaluating prediction for a binary classification task, we report area under curve (AUC) on test set. For the multiclass DAG datasets, we report test set accuracy.When evaluating prediction performance for a regression task, we use Root Mean Squared Error (RMSE) on test set.  For the evaluation of estimated dependency structure on simulated cases (in which we know the true dependency graphs), we report average AUC
to compare estimated graph probability (or interaction score from baselines) with the true binary graphs. For the cases, where we do not know the ground truth dependency graph, we use pairwise classification metric to show the utility of the learnt graphs.

\paragraph{Baselines: } (1) \textbf{QDA: } We use QDA as one baseline on simulation cases as it is close to ground truth for classifying Gaussian Datasets with distinct covariance matrices. (2) \textbf{MLP:} we compare against multi-layer perceptron networks (MLP). (3) \textbf{GAT-FC: } We compare against a GAT without graph training (GAT-FC). Here we use a Fully Connected graph to GAT. For Graph Recovery, we compare against Neural Interaction Detection\cite{tsang2017detecting}(NID).

\paragraph{Variations:} We have the following versions for \methodName:
\begin{itemize}[noitemsep,topsep=0pt]
    \item {\LmethodName}:  Multiple graphs variation of \methodName.
    \item {\YmethodName}: We add $y$ to the graph for the structure learner. Thus, this variation effectively learns a graph between the nodes $V \in \{\boldsymbol{x}_i, y\}\forall i \in \{1,\dots, p\}$. In the task learner, we do not use the true $y$, instead use the $CLS$ node as a proxy for $y$. 
    \item {\LYmethodName}: This variation combines the above two methods, hence learns a class specific graph between 
\end{itemize}

\begin{table*}[t]
\small
   \centering
    \begin{tabular}{|l|c|c|c|c|c|c|c|}\hline

           \makecell{Models, Baselines, \\  Ablations and Variations} & \makecell{Prediction-AUC \\ ($p=5$)} & \makecell{Prediction-AUC \\ ($p=10$)} & \makecell{Prediction-AUC \\ ($p=20$)} & \makecell{Graph-AUC \\ ($p=5$)} & \makecell{Graph-AUC \\ ($p=10$)} & \makecell{Graph-AUC \\ ($p=20$)} \\\hline\hline
        {\methodName} & 0.7132 & 0.7145 &  \textbf{0.8491} & \textbf{1.0} & 0.9979 & 0.9957 \\\hline
        {\methodNameNSparse}  & 0.7139 & 0.7162 &  0.8479 &  0.4667
 & 0.4713 &  0.4462\\\hline
        {\methodNameNS}  & 0.7144 & \textbf{0.7192} & 0.8471 & 0.4521 & 0.4602 &  0.5297
        \\\hline
        M-{\methodName} & \textbf{0.7147} & 0.7142  & 0.8467 & \textbf{1.0}  & \textbf{1.0} & \textbf{1.0} \\\hline
        M-{\methodNameNSparse} &  0.7135 & 0.7153 & 0.8443 & 0.3396
& 0.5680
  & 0.5912 \\\hline
        M-{\methodNameNS} &  0.7139 & 0.7165 & 0.8453 & 0.6021
 & 0.4567
 & 0.4671\\\hline
  
        NID &  NA & NA & NA & 0.6250 & 0.6526 & 0.6300 \\\hline
       
          GAT-FC & 0.7137 & 0.7051 & 0.8489 & NA & NA & NA \\\hline
         MLP & 0.7161 & 0.7193 & \textbf{0.8548} & NA & NA & NA \\\hline
        QDA & \textbf{0.7178} & \textbf{0.7252} & 0.8215 & NA & NA & NA  \\\hline 
    \end{tabular}
    \caption{Classification Results Area under Curve (AUC) on test data and Evaluation on Graph Estimations  for simulation datasets averaged across $5$ random seeds. %
    }
    \label{tab:shared_sim}
\end{table*}

 \subsection{Simulated: Gaussian MRF-based Classification Datasets}

First, we try to answer the question: can \methodName learn to distinguish two Gaussian distributions while simultaneously learning the underlying conditional independence relationships, represented by the precision matrix?

{\bf Data Generation}: We consider a simple binary classification task. For each class, we use a multivariate normal distribution to generate simulated samples. For a multivariate normal distribution, the precision matrix $\boldsymbol{\Omega}$, inverse of the covariance matrix, is representative of the conditional dependency graph. The values indicate partial correlation between two variables, while the zero entries represent conditional independence given all other variables (therefore no edge in the graph). Specifically,
$\boldsymbol{\Omega}_{ij}$=0 if and only if $\boldsymbol{X}_i$ and $\boldsymbol{X}_j$ are conditionally independent given all other coordinates of $\boldsymbol{X}$. 

In detail, we simulate two dependency graphs being sampled as Erdos Renyi Graphs: matrices $\boldsymbol{\Delta}$ and  $\boldsymbol{R}_I \in \mathbb{R}^{p \times p}$,   with probability $p_d$ and $p_i$ respectively. 
Then we generate data from two classes $A$ and $B$ using the following equations:
$\boldsymbol{\Omega}_A = \boldsymbol{\Delta} + \boldsymbol{R}_I + \delta_d I $, and  $\boldsymbol{\Omega}_B =  \boldsymbol{R}_I+\delta_d I$. $\delta_c$ and $\delta_d$ are selected large enough to guarantee positive definiteness.  We generate Gaussian samples using $\boldsymbol{X}_A \sim N(0,\boldsymbol{\Omega}_A^{-1})$ and $\boldsymbol{X}_B \sim N(0,\boldsymbol{\Omega}_B^{-1} )$. 
\paragraph{Ablations: } To understand how each component impact \methodName, we explore the following: 
\begin{itemize}[noitemsep,topsep=0pt]
    \item \methodName:  Basic version with one graph.
    \item {\methodNameNS}: \methodName is trained without self-supervision loss. Concretely, $\lambda_{struct}=0$. We only use the task loss with sparsity to train the graph. 

    \item {\methodNameNSparse}: In this variation, \methodName is trained without self-supervision loss  and sparsity regularization is also removed. Concretely, $\lambda_{struct}=0$, $\lambda_{sparse}=0$.
    
    \item {\LmethodNameNS}: M-\methodName is trained without self-supervision loss. .i.e.,  $\lambda_{struct}=0$.
    \item {\LmethodNameNSparse}:  M-\methodName is trained without self-supervision loss. and sparsity regularization is also removed in {\LmethodNameNS} .i.e.,  $\lambda_{struct}=0$, $\lambda_{sparse}=0$.
\end{itemize}

\begin{figure*}[t]
\begin{center}
\setlength\tabcolsep{0.3pt} 
\begin{tabular}{cc}
\includegraphics[scale=0.3]{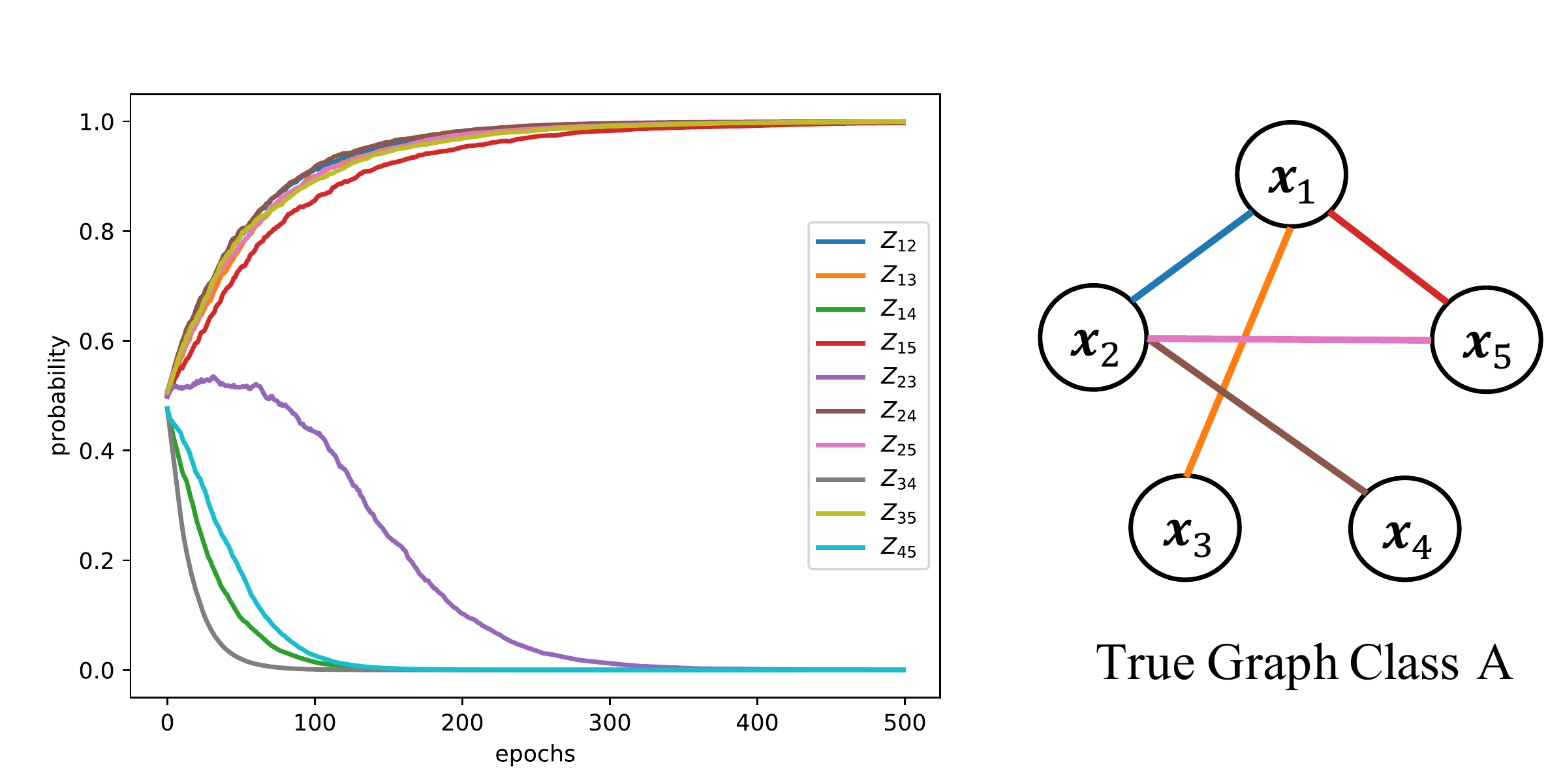}&
\includegraphics[scale=0.3]{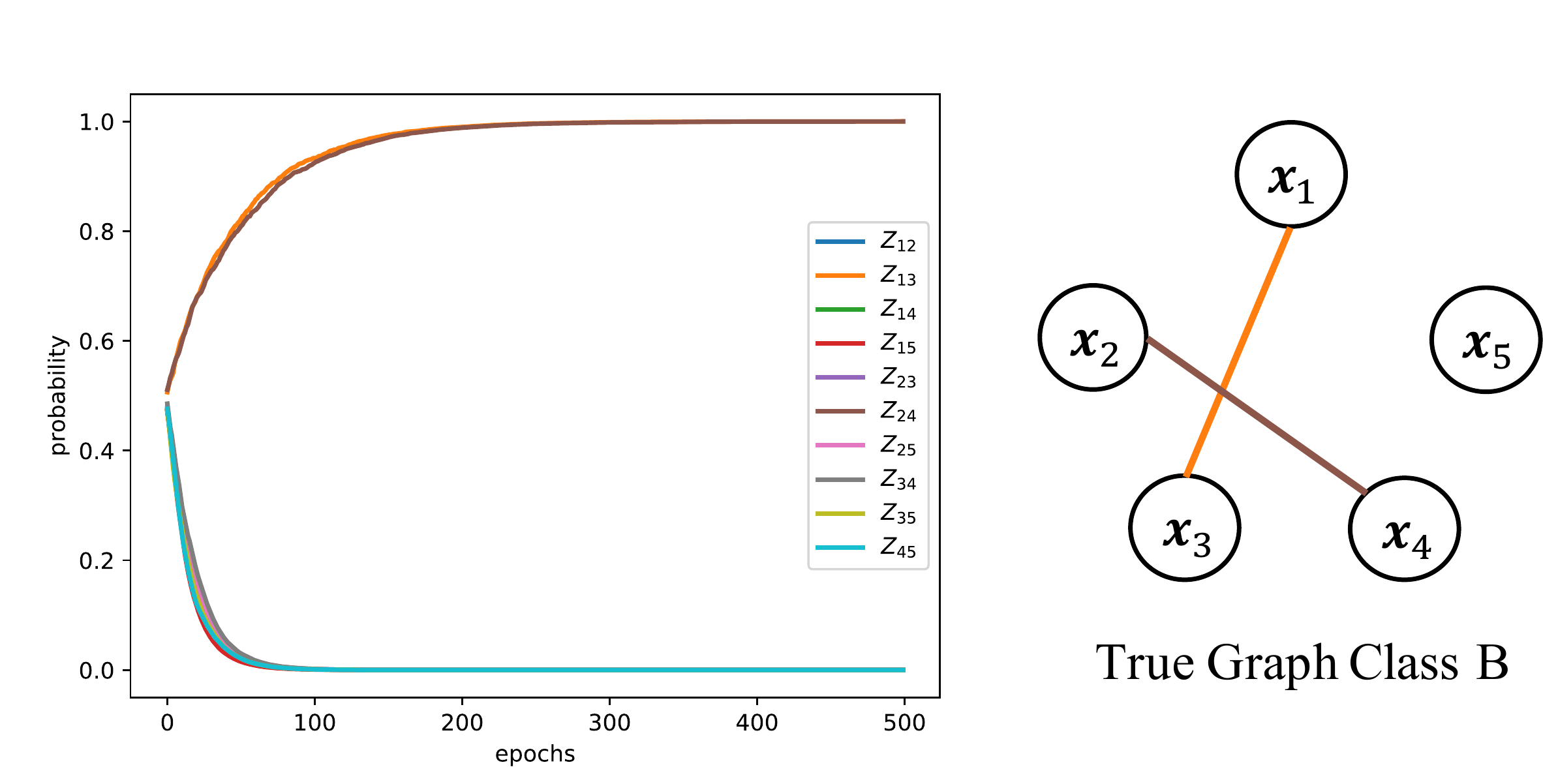}\\
\end{tabular}
\end{center}
\vspace{-0.1in}
\caption{On Simulation Data: We show convergence during training
of the Graph parameters $\sigma(\boldsymbol{\gamma}_{ij})$ indicating the probability of the edge in the graph structure. We show the ground truth graphs True Graph Class A and True Graph Class B adjacent to the convergence graphs. The true edges converge to probability $1.0$ and the true non-edges converge to probability $0.0$.%
\label{fig:convergence}
}
\end{figure*}

{\bf Experiment Setup and Hyperparameter Tuning}: We consider three cases: $p=5$, $p=10$ and $p=20$. For $p=5$, we consider the graphs shown in Figure~\ref{fig:convergence}. For $p=10$, we consider $p_d=0.3$ and $p_i=0.2$. For $p=20$, we consider $p_d=0.3$ and $p_i=0.1$. We use training samples $n=40000$, and validation and test samples, each as $8000$.  In Table~\ref{tab:shared_sim}, we present the classification results of \methodName and variations. We search over the best hyperparameter setting for the task learner GAT over the following combinations of $(e_H, t_H, t_L, nheads, t_D)$ as [(16,32,2,4,0.0),
          (16,64,2,4,0.0),
          (16,32,4,4,0.0),
          (64,32,2,4,0.0),
          (64,32,4,4,0.0),
          (16,64,2,4,0.5),
          (16,32,4,4,0.5),
          (64,32,2,4,0.5),
          (64,32,4,4,0.5].

We report the results averaged across $5$ random seeds.  Table~\ref{tab:shared_sim} shows our classification results(AUC) as well as our graph recovery results(AUC wrt the true graph) on Gaussian Datasets for all settings.  We use bold text to represent the best variation from our model and from the baselines. We report the  standard deviation across the seeds in the Appendix. Figure~\ref{fig:convergence} shows  that the Bernoulli parameters $\sigma(\boldsymbol{\gamma})$ converge, with edges converging to probability $1.0$ and non edges to probability $0.0$.

\vspace{-2mm}

\subsection{Real World Datasets:Directed Acyclic Graph Datasets}
Here, we try to answer the question if \methodName can learn interpretable and meaningful graphs on well-known classification datasets  from a datasets from the University of California at Irvine (UCI) repository. 
\paragraph{Datasets}: Table~\ref{tab:dag_splits} provides a summary of the datasets and corresponding data splits. 
\begin{table}[h]
    \centering
    \begin{tabular}{|c|c|c |c|}\hline
         Dataset & $n_{train}$ / $n_{valid}$ / $n_{test}$ & $p$  & $C$ \\ \hline
         pima & 621/70/77 & 6 & 2\\\hline
         vehicle & 684/77/85 & 18 &4 \\\hline
         satimage & 3991/444/2000 & 36 &6 \\\hline
         waveform & 4050/450/500 & 21 &3 \\\hline
         housing & 16718/1858/2064 & 8 & regression\\ \hline 
         heart & 202/45/51 & 13 & 2\\ \hline 
    \end{tabular}
    \caption{Description of DAG datasets used in the experiments: $p$ represents the number of input features, $C$ represents the number of classes. }
    \label{tab:dag_splits}
\end{table}
\begin{table*}[]
    \centering
    \small
    \begin{tabular}{|c|c|c|c|c|c|c|c|}\hline
         Dataset  & \methodName  & \YmethodName & \LmethodName & \LYmethodName & \methodName-FC & MLP & BNC\cite{friedman1997bayesian} \\\hline\hline
       vehicle  &  0.862 & 0.860 & {\bf 0.882} & 0.859  & 0.812 & 0.847 & 0.696 \\\hline
       pima  & 0.7800 & {\bf 0.808} &  0.795 &  0.714 & 0.779  & 0.766  & 0.755\\\hline
       waveform  & 0.882   & 0.881 & 0.861 & 0.875  & 0.874 & {\bf 0.884 } &  0.784 \\\hline 
       satimage  & 0.913 & 0.905 & 0.911 & 0.902 & 0.908 & 0.913  &  0.872 \\\hline 
       housing & {\bf 0.497} & 0.591  &NA &  NA& 0.504 &  0.503 & NA\\\hline
       heart & 0.9276  &  0.930 &  0.962 & 0.920 &0.924 & {\bf 0.973} & 0.833\\\hline
    \end{tabular}
    \caption{Test Accuracy for DAG datasets}
    \label{tab:dag_acc}
\end{table*}

\paragraph{Hyperparameters: } For all our models, we use a $\lambda_{sparse}=0.001$, $\lambda_{dag}=0.01$. we use $\lambda_{struct}=1.0$. We pretrain both the structure learner and the task learner using a fully connected graph. We search over the best hyperparameter setting for the task learner GAT over the following combinations of $(e_H, t_H, t_L, nheads, t_D)$ as [(16,32,2,4,0.0),
          (16,64,2,4,0.0),
          (16,32,4,4,0.0),
          (64,32,2,4,0.0),
          (64,32,4,4,0.0),
          (16,64,2,4,0.5),
          (16,32,4,4,0.5),
          (64,32,2,4,0.5),
          (64,32,4,4,0.5]. We use  Adam Optimizer with a learning rate of 0.001. We report the average test performance for the architecture with best validation performance. For MLP baseline model, we search over the following combinations of $t_H \in \{32,64,128,256\}$, $t_L\in \{2,4\}$ and $t_D\in \{0.0,0.5\}$. We show our dataset splits in Table~\ref{tab:dag_splits}. For the housing dataset, which is a regression dataset, we do not use the concatenated $\boldsymbol{x}$ to each node feature $\boldsymbol{x}_i$, based on validation performance. 
          
\begin{figure}[tbh]
    \centering
    \includegraphics[width=\linewidth]{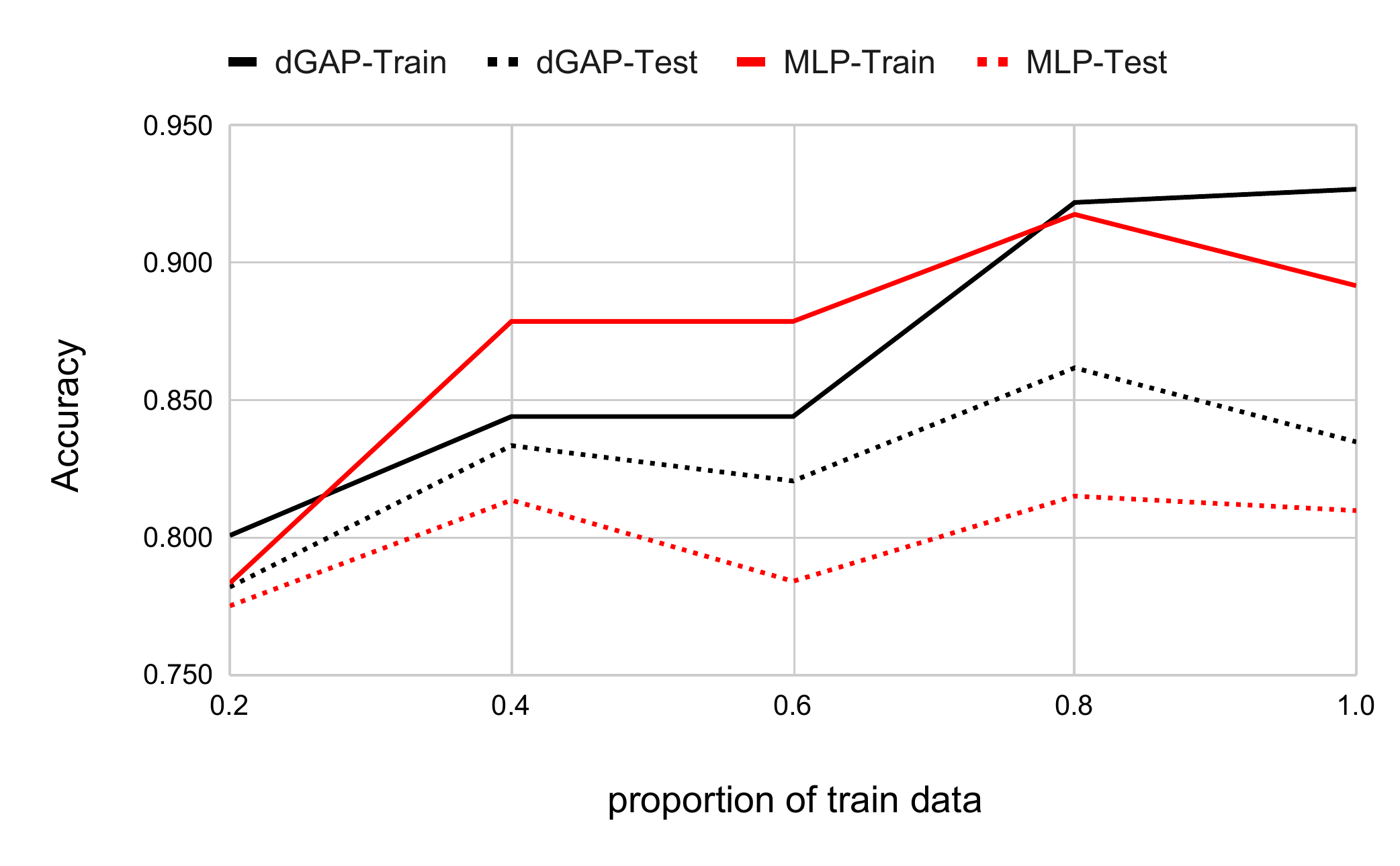}
    \caption{Data Efficiency on Vehicle UCI dataset}
    \label{fig:vehicle_data_eff}
\end{figure}

\begin{figure}[tbh]
\begin{tabular}{cc}
  \includegraphics[width=0.4\linewidth]{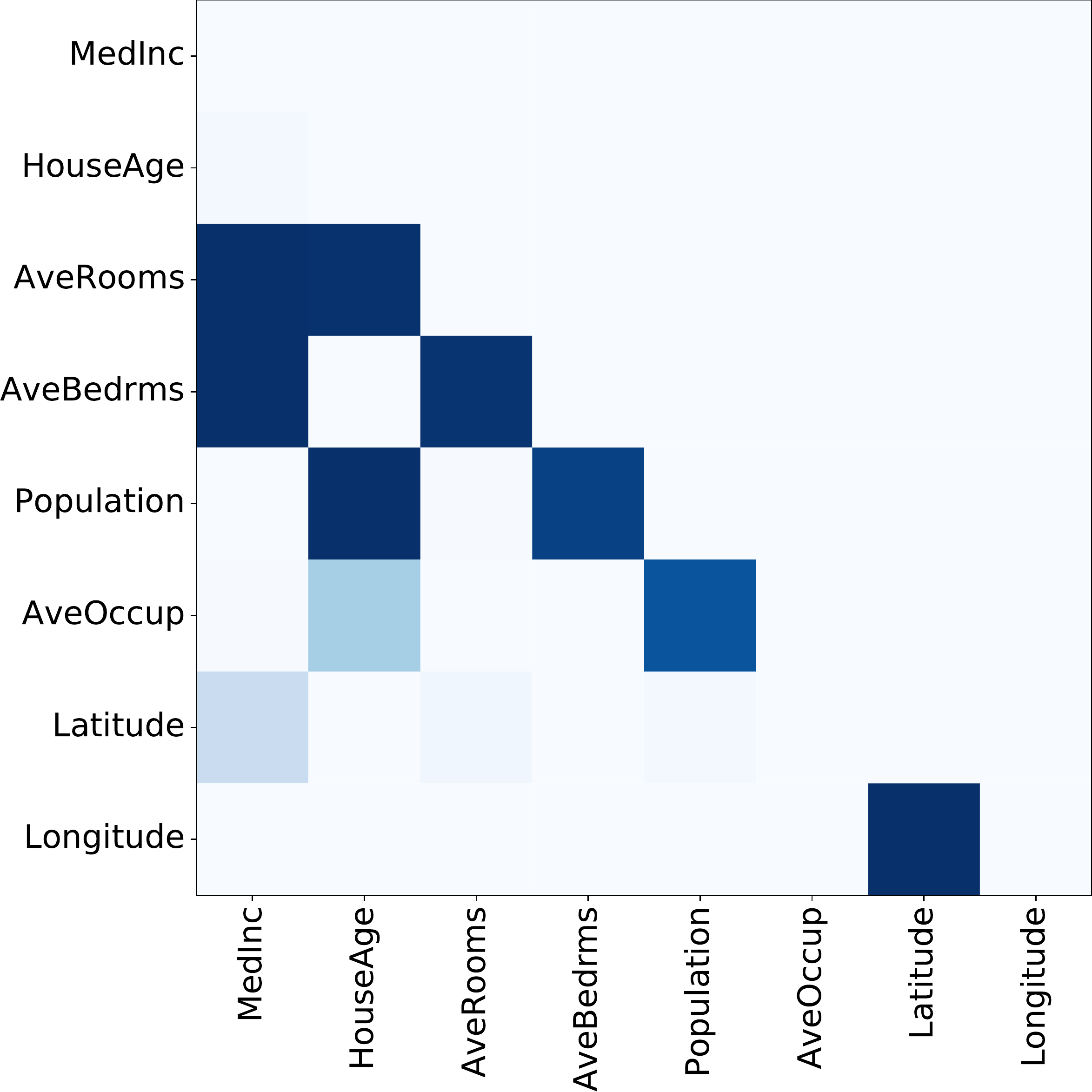} &   \includegraphics[width=0.4\linewidth]{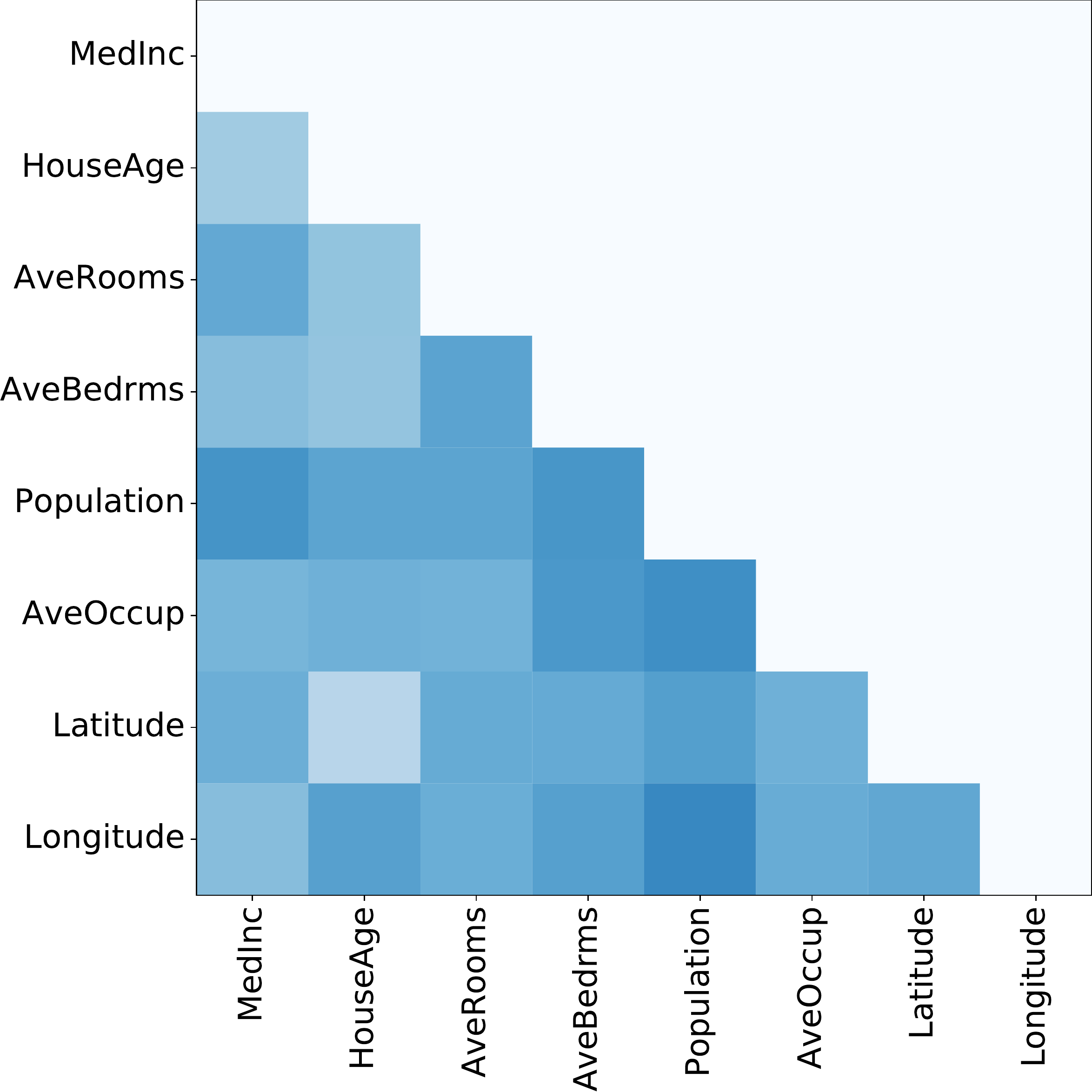} \\
(a)  & (b)   \\
 [6pt]

\end{tabular}

\caption{Heatmaps of pairwise interaction strengths learnt by {\methodName}  on the Cal-Housing data: (a) Graph learnt by {\methodName}. Our method is able to successfully learn a relationship between latitude and longitude; (b) Graph learnt by {\methodNameNSparse}. While some interactions are shared with (a), without structure loss we cannot recover  a strong relationship between latitude and longitude. 
}
\label{fig:nid_cal}
\end{figure}

\paragraph{Results:}~We show our results in Table~\ref{tab:dag_acc}. \methodName achieves a higher test accuracy compared to the baselines in 4/6 datasets.\methodName also provides interpretability. 
Figure~\ref{fig:nid_cal}(a) shows that our method is able to successfully learn a relationship between latitude and longitude which is important for house value prediction.  We also show the case where we use solely the task loss to learn a graph in Figure~\ref{fig:nid_cal}(b). While some interactions are similar to (a), without structure loss we cannot recover a strong relationship between latitude and longitude. We present the visualization results from other datasets in the appendix.
\paragraph{Data Efficiency: } ~We subsampled the training data from Vehicle dataset, with the same validation and test split and
retrained each model. We use the model with the best validation performance on the entire training data (used in the previous section). In Figure~\ref{fig:vehicle_data_eff}, we observe that \methodName is more data efficient compared to the baseline.

\begin{figure}[hbt]
\begin{tabular}{ccc}
  \includegraphics[width=0.3\linewidth]{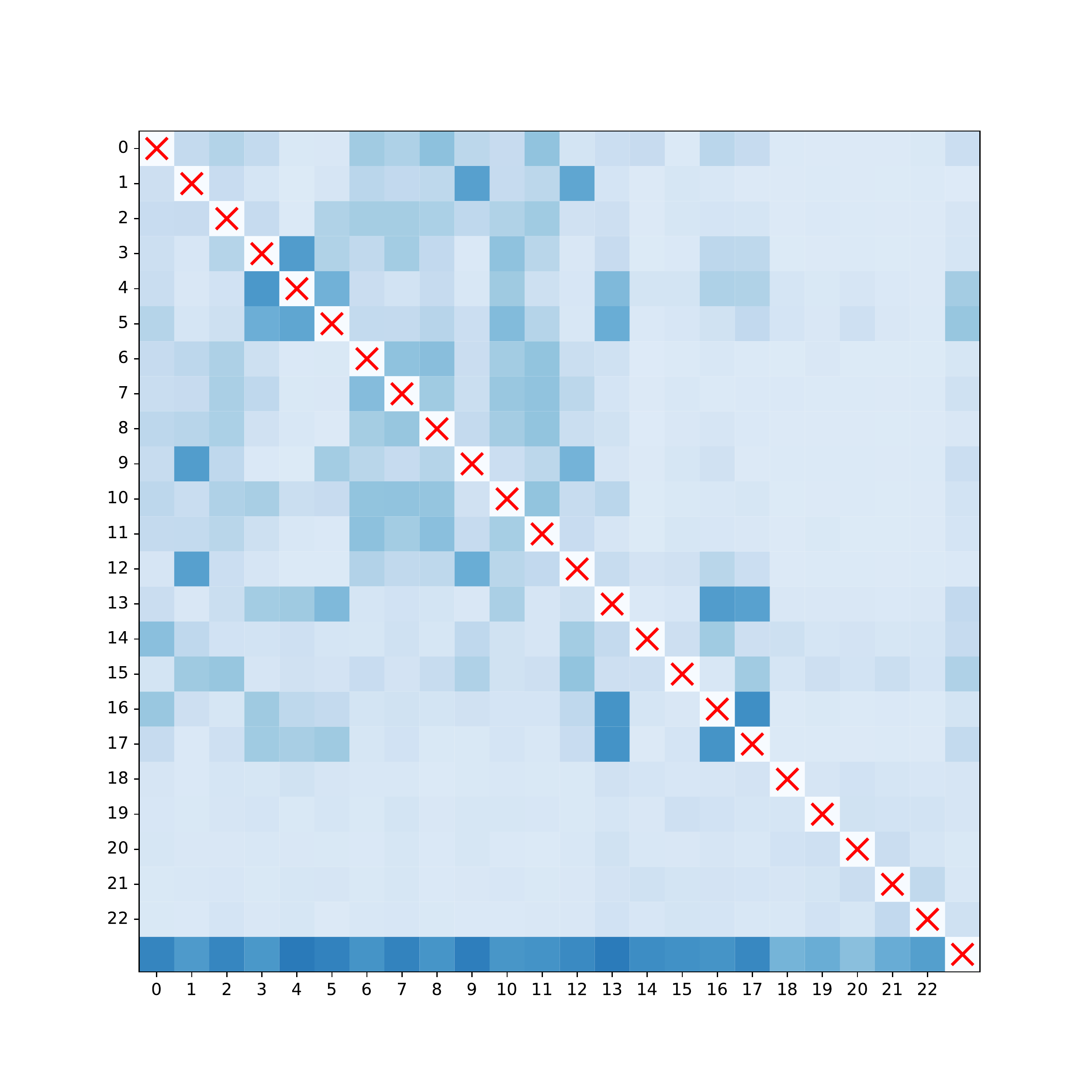} &   \includegraphics[width=0.3\linewidth]{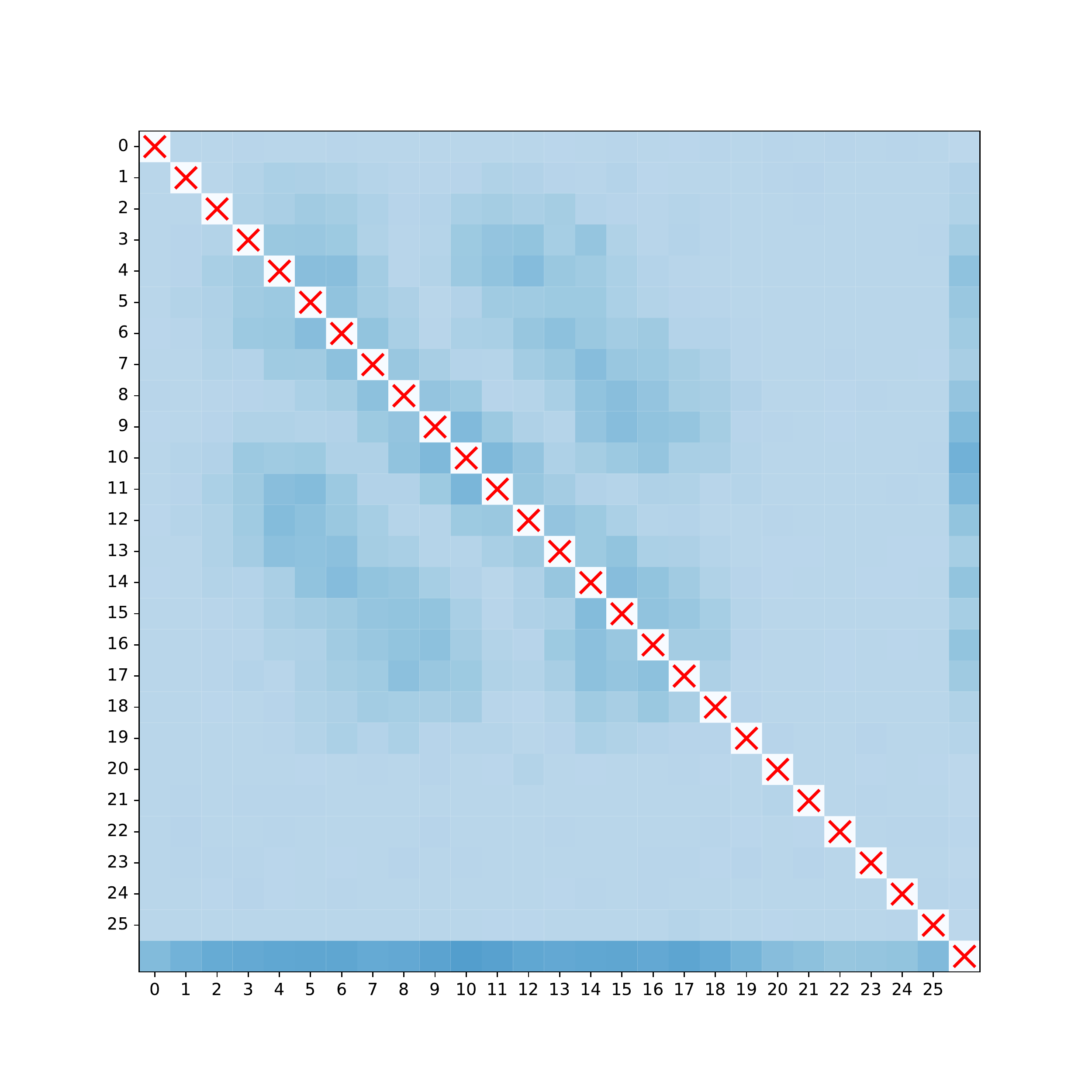} &

 \includegraphics[width=0.3\linewidth]{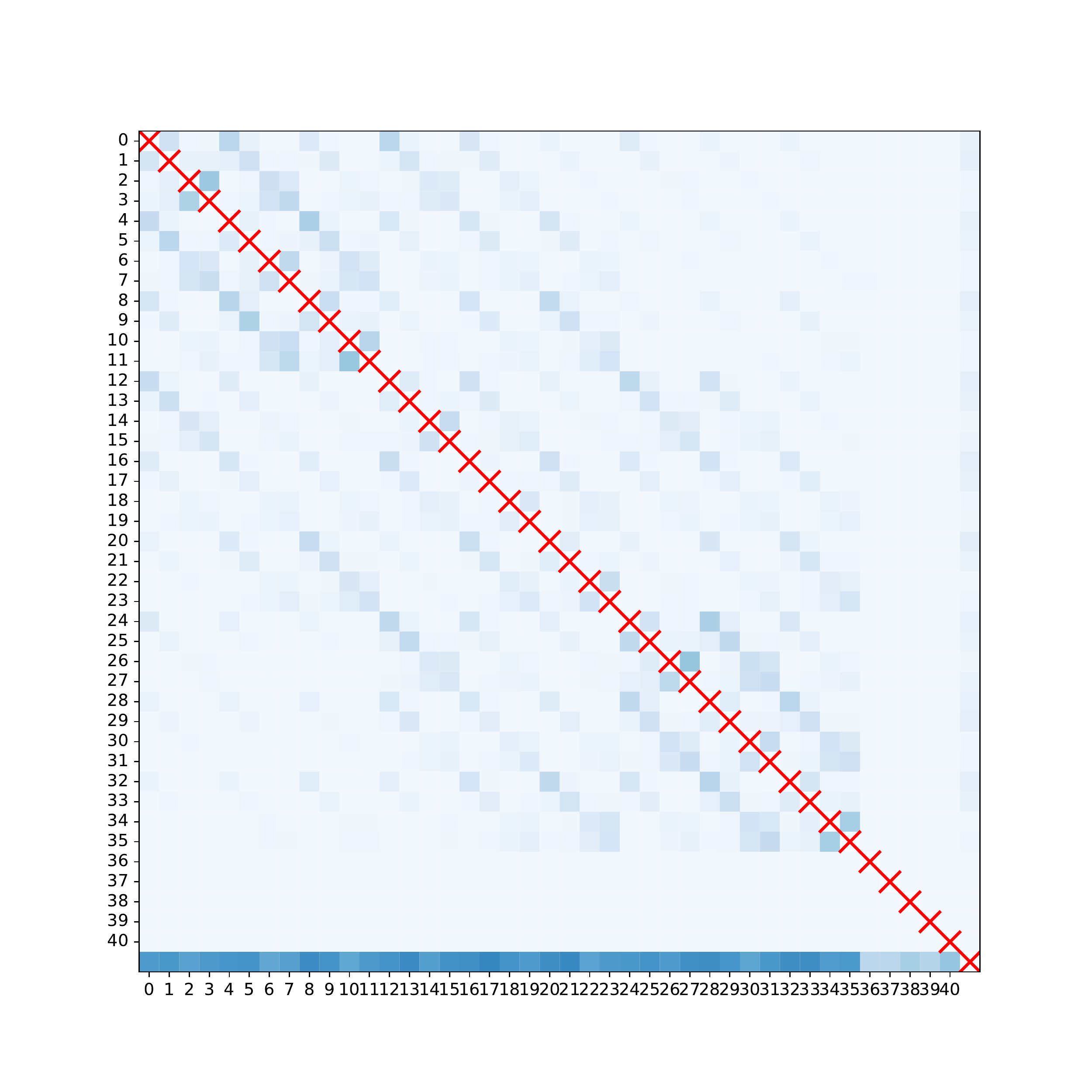} \\
(a) Vehicle & (b)  Waveform & (c) Satimage  \\[6pt]

\end{tabular}

\caption{When with Noisy Features: Learnt Graph on DAG datasets with noisy features,  the last 5 features are noisy and dGAP-Y successfully learns a lower probability of the edges from the features to label $y$, denoted by the last row(the noisy features are the 5 features numbered 18-22).}
\label{fig:noise_viz}
\end{figure}

\paragraph{Noisy Features:}~\methodName-Y includes label $y$ as part of the learnt graph $\boldsymbol{Z}$. This enables the model to select features that are informative for prediction of $y$. We add $5$ `noise' features drawn from a normal distribution to three DAG datasets.  We show in Table~\ref{tab:noise} that dGAP achieves a higher test accuracy compared to baseline MLP. We also show that \methodName-Y learns a lower probability of the edges  corresponding to noisy features  $\boldsymbol{x}_n$ to $y$. We visualize the learnt graphs by \methodName-Y in Figure~\ref{fig:noise_viz}. We show that \methodName-Y successfully assigns a lower probability to the noisy features.

\begin{table}[tbh]
\centering
\begin{tabular}{|c|c|c|}\hline
dataset   & dGAP-Y-noise & MLP-noise \\\hline
vehicle   & {\bf 0.837 }       & 0.788     \\\hline
waveform &      {\bf 0.878  }      & 0.86      \\\hline
satimage &      {\bf 0.899 }        & 0.887  \\\hline  
\end{tabular}

\caption{Test Accuracy of \methodName-Y with noisy features.}
\label{tab:noise}
\end{table}

\begin{figure}[tbh]
\begin{tabular}{cc}
  \includegraphics[width=0.4\linewidth]{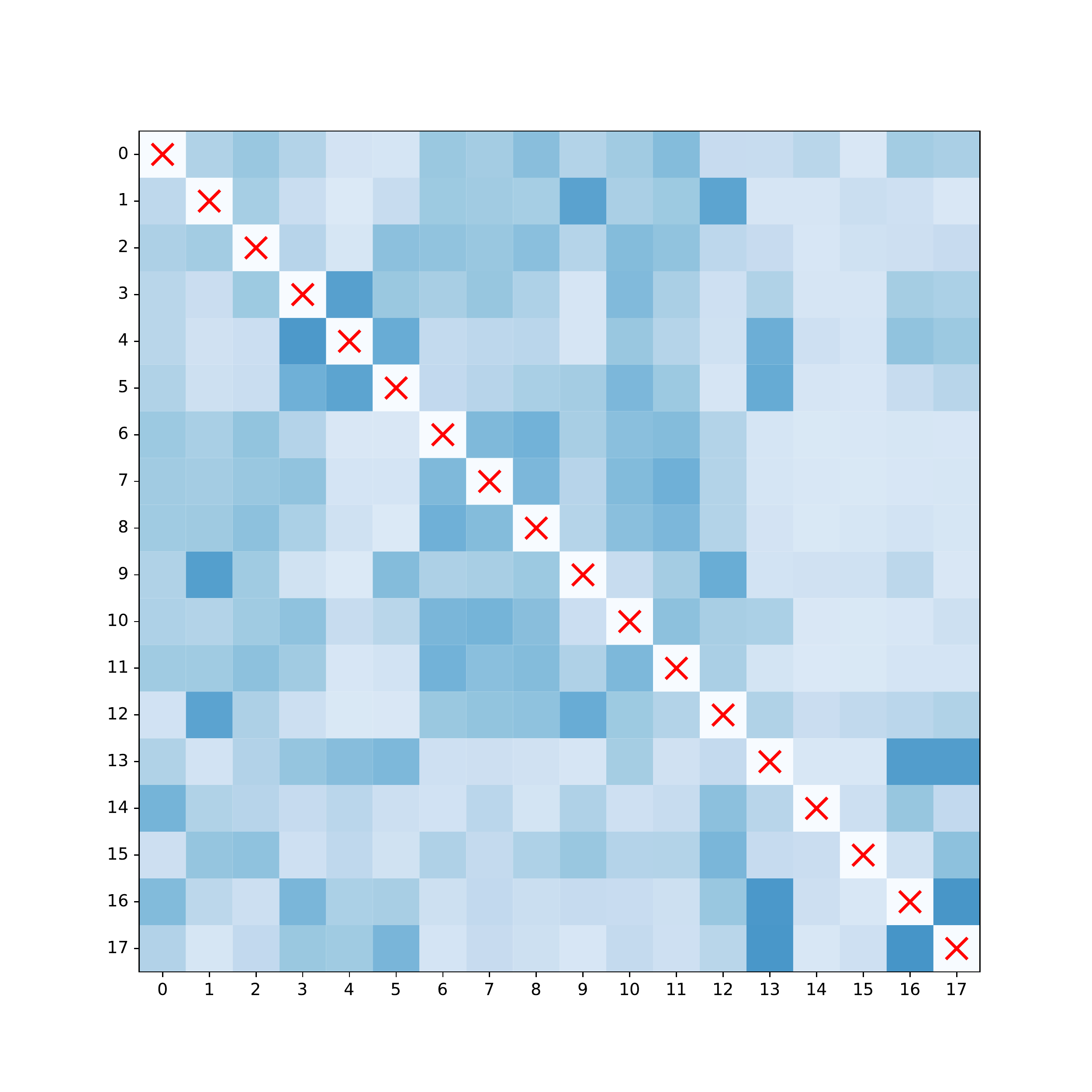} &   \includegraphics[width=0.4\linewidth]{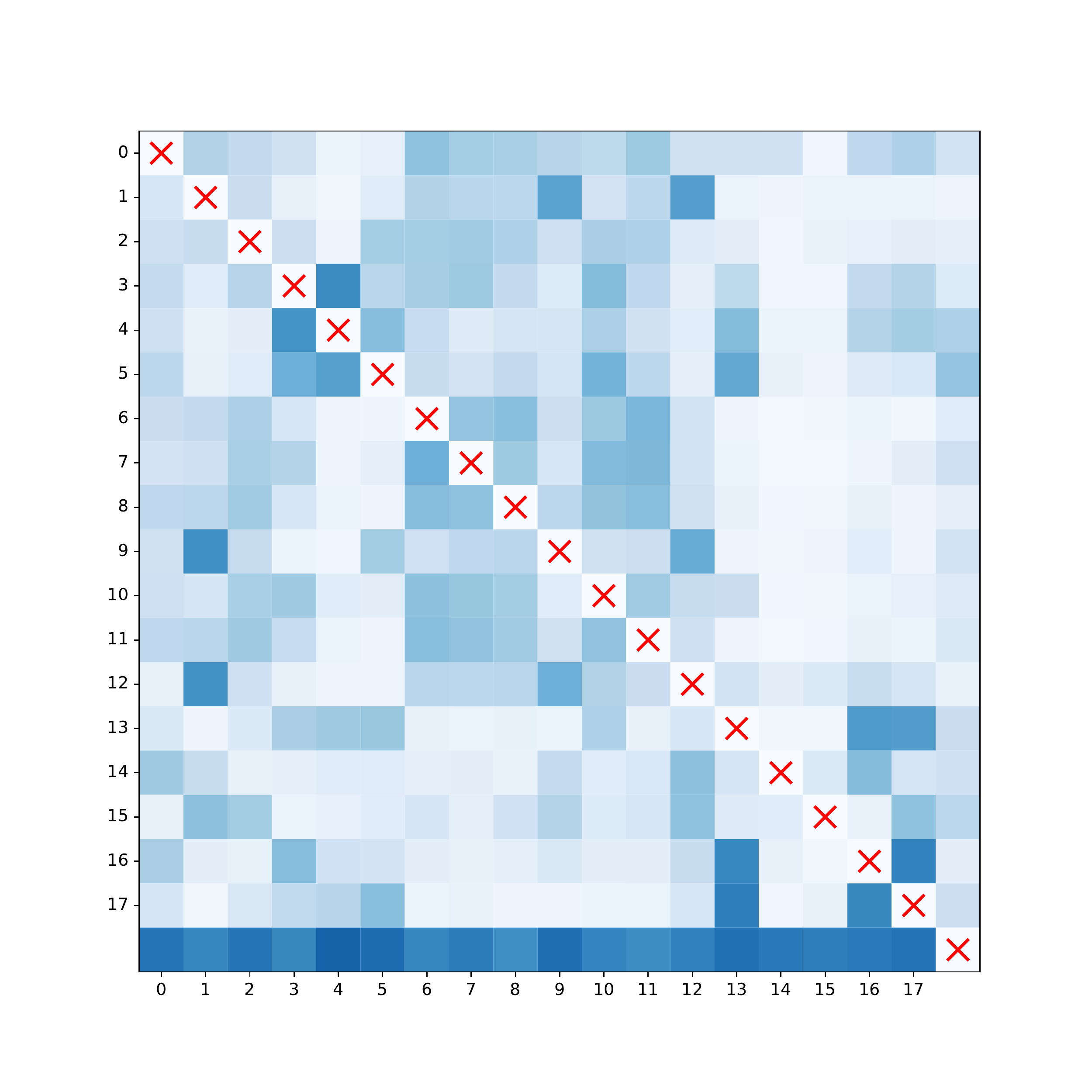} \\
(a) Vehicle \methodName & (b)  Vehicle \YmethodName \\
 \includegraphics[width=0.4\linewidth]{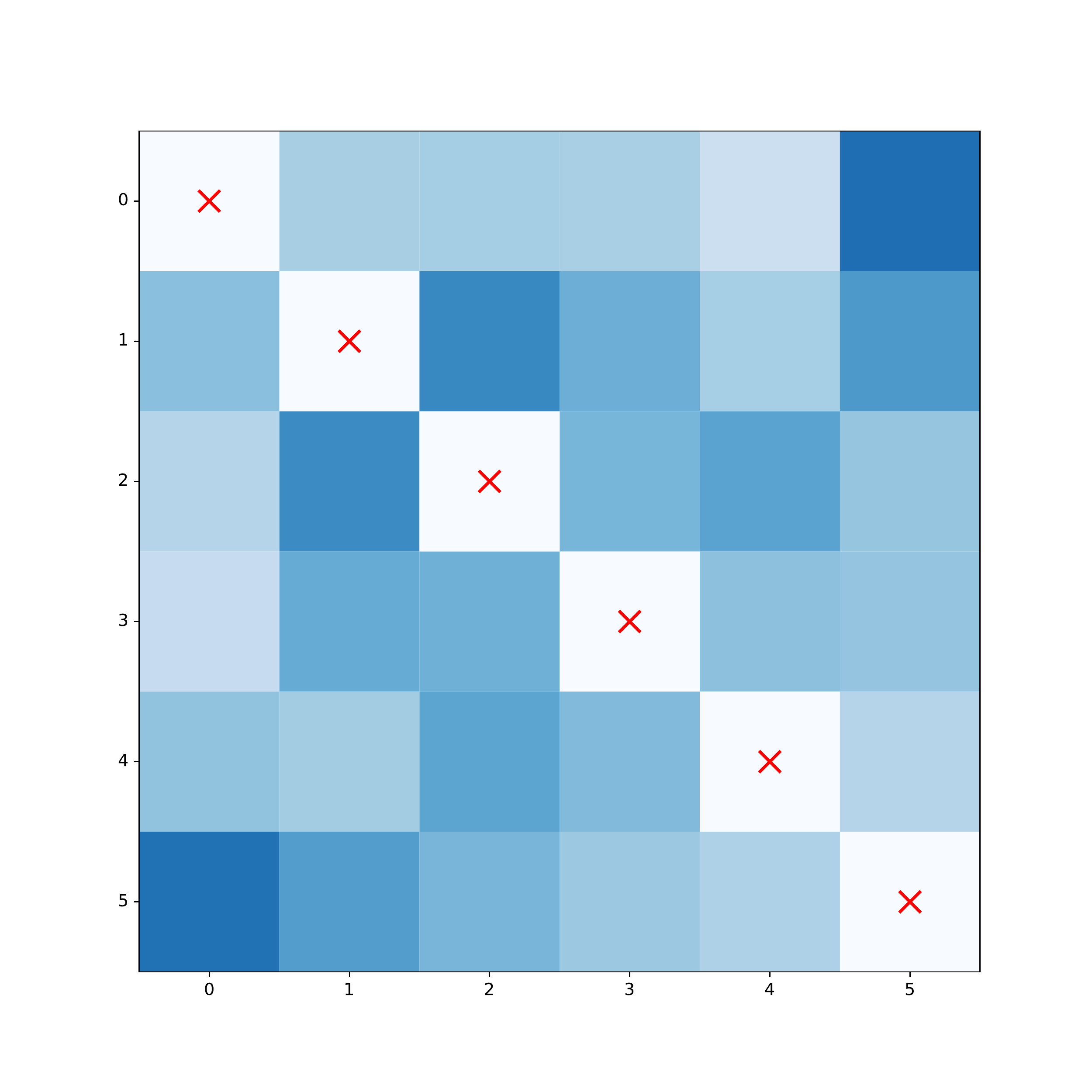} &
 \includegraphics[width=0.4\linewidth]{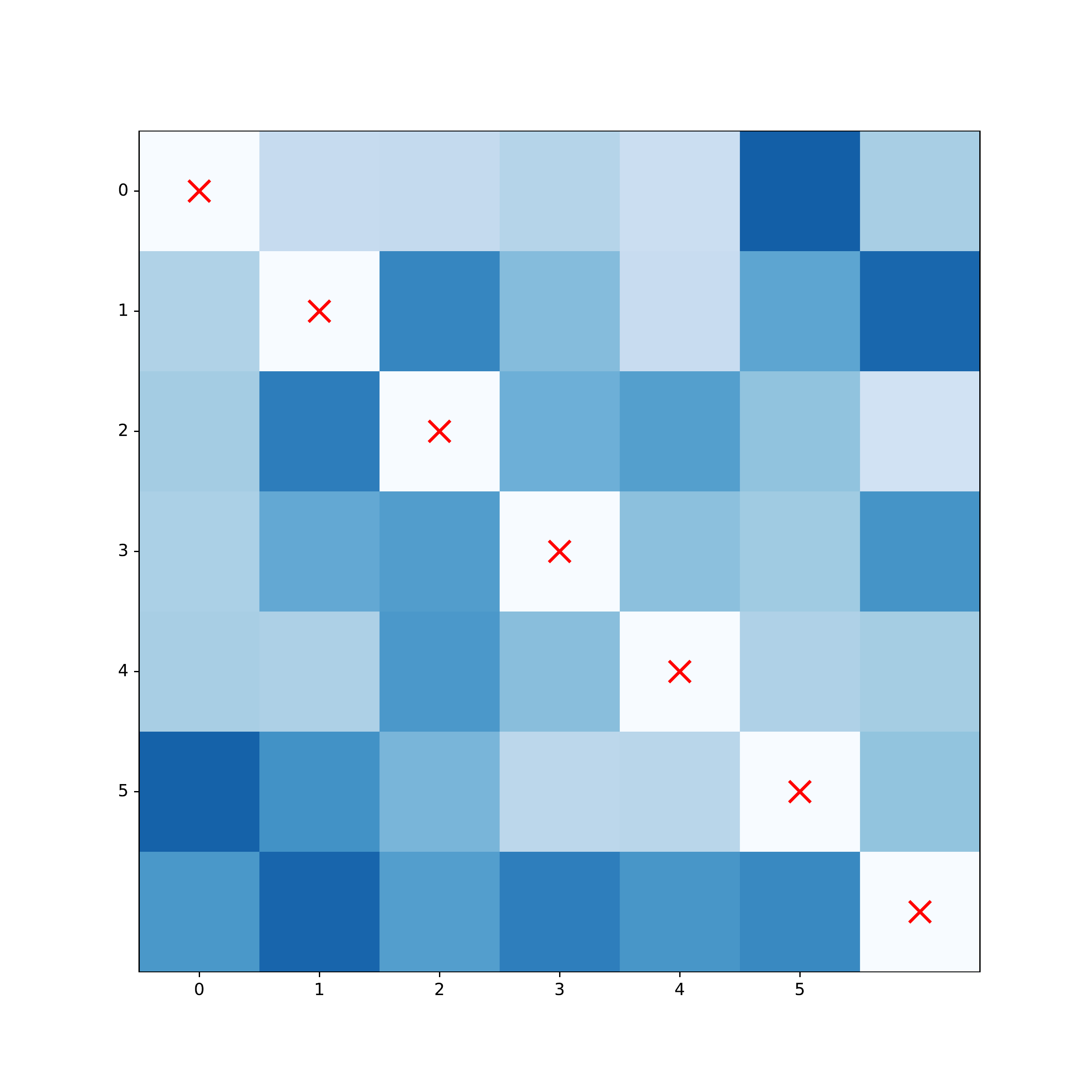} \\
 (c) PIMA \methodName & (d) PIMA \YmethodName.\\[6pt]

\end{tabular}

\caption{Visualization of learnt graph on vehicle and pima dataset.}
\label{fig:dag_viz}
\end{figure}

\subsection{CUB Dataset}
We use the Caltech-UCSD Birds-200-2011 (CUB) dataset \cite{welinder2010caltech}, which comprises $n = 11,788$ bird photographs. The labels correspond to $112$ labels.  As nodes, we use $k = 112$ binary bird attributes representing wing color, beak shape, etc. We use the same labeling scheme for the concepts as \cite{koh2020concept}. These correspond to the class-level concepts, where each class of birds has the same concepts. 

Similar to the Concept Bottleneck models, we resize one of the layers to match the number of concepts and use them to predict the concepts by augmenting the model with an auxiliary concept prediction loss.  Given that the concepts have high concept prediction accuracy and are valid for the task at hand, we augment the Concept BottleNeck Model with a dGAP model, that can simultaneously learn the network between concepts. We use the same $x\rightarrow c $ model as \cite{koh2020concept}. For the $c\rightarrow y$ model, we replace the MLP used in \cite{koh2020concept} with \methodName. We show our results in Table~\ref{tab:cub}. \methodName achieves lower test error compared to \cite{koh2020concept}.

\begin{table}[H]
    \centering
    \begin{tabular}{|c|c|} \hline
    Model  &  Test Error \\ \hline
        Joint-ConceptBottleNeck-MLP\cite{koh2020concept} &  0.199 \\\hline
       Joint-ConceptBottleNeck-dGAP  & { \bf 0.182} \\ \hline
    \end{tabular}
    \caption{Test Error on CUB Dataset}
    \label{tab:cub}
\end{table}

\section{Conclusion}

This paper proposes an end-to-end deep learning framework, \methodName, to learn graph-structured representations that help in predictions. We can view \methodName regarding two different ways. (1) We can view \methodName as an approach for discovering task-oriented structured representations that are critical components contributing towards human-like cognitive data modeling. 
\methodName is an explicit neural architecture to identify dependency edges and use edges as part of the representations to enable prediction goals. (2) We can also view \methodName as being ``modular" interpretable~\cite{murdoch2019definitions}.  ``Modular" interpretable\cite{murdoch2019definitions} is defined as a model in which a meaningful portion of its prediction-making process can be interpreted independently. Empirically, we show,  on multiple simulated and real-world datasets, the structure graph representations learned by \methodName are useful for knowledge communication and for reasoning using knowledge to generate new decisions.

\bibliographystyle{abbrv}
\bibliography{deepstructure}

\begin{thebibliography}{10}

\bibitem{besag1977efficiency}
J.~Besag.
\newblock Efficiency of pseudolikelihood estimation for simple gaussian fields.
\newblock {\em Biometrika}, pages 616--618, 1977.

\bibitem{cui2019recovering}
T.~Cui, P.~Marttinen, and S.~Kaski.
\newblock Recovering pairwise interactions using neural networks.
\newblock {\em arXiv preprint arXiv:1901.08361}, 2019.

\bibitem{devlin2018bert}
J.~Devlin, M.-W. Chang, K.~Lee, and K.~Toutanova.
\newblock Bert: Pre-training of deep bidirectional transformers for language
  understanding.
\newblock {\em arXiv preprint arXiv:1810.04805}, 2018.

\bibitem{franceschi2019learning}
L.~Franceschi, M.~Niepert, M.~Pontil, and X.~He.
\newblock Learning discrete structures for graph neural networks.
\newblock {\em arXiv preprint arXiv:1903.11960}, 2019.

\bibitem{friedman2008predictive}
J.~H. Friedman, B.~E. Popescu, et~al.
\newblock Predictive learning via rule ensembles.
\newblock {\em The Annals of Applied Statistics}, 2(3):916--954, 2008.

\bibitem{friedman1997bayesian}
N.~Friedman, D.~Geiger, and M.~Goldszmidt.
\newblock Bayesian network classifiers.
\newblock {\em Machine learning}, 29(2-3):131--163, 1997.

\bibitem{halford2010relational}
G.~S. Halford, W.~H. Wilson, and S.~Phillips.
\newblock Relational knowledge: the foundation of higher cognition.
\newblock {\em Trends in cognitive sciences}, 14(11):497--505, 2010.

\bibitem{jang2016categorical}
E.~Jang, S.~Gu, and B.~Poole.
\newblock Categorical reparameterization with gumbel-softmax.
\newblock {\em arXiv preprint arXiv:1611.01144}, 2016.

\bibitem{janizek2020explaining}
J.~D. Janizek, P.~Sturmfels, and S.-I. Lee.
\newblock Explaining explanations: Axiomatic feature interactions for deep
  networks, 2020.

\bibitem{ke2019learning}
N.~R. Ke, O.~Bilaniuk, A.~Goyal, S.~Bauer, H.~Larochelle, C.~Pal, and
  Y.~Bengio.
\newblock Learning neural causal models from unknown interventions.
\newblock {\em arXiv preprint arXiv:1910.01075}, 2019.

\bibitem{kipf2018neural}
T.~Kipf, E.~Fetaya, K.-C. Wang, M.~Welling, and R.~Zemel.
\newblock Neural relational inference for interacting systems.
\newblock {\em arXiv preprint arXiv:1802.04687}, 2018.

\bibitem{koh2020concept}
P.~W. Koh, T.~Nguyen, Y.~S. Tang, S.~Mussmann, E.~Pierson, B.~Kim, and
  P.~Liang.
\newblock Concept bottleneck models.
\newblock {\em arXiv preprint arXiv:2007.04612}, 2020.

\bibitem{lachapelle2019gradient}
S.~Lachapelle, P.~Brouillard, T.~Deleu, and S.~Lacoste-Julien.
\newblock Gradient-based neural dag learning.
\newblock {\em arXiv preprint arXiv:1906.02226}, 2019.

\bibitem{lundberg2018consistent}
S.~M. Lundberg, G.~G. Erion, and S.-I. Lee.
\newblock Consistent individualized feature attribution for tree ensembles.
\newblock {\em arXiv preprint arXiv:1802.03888}, 2018.

\bibitem{maddison2016concrete}
C.~J. Maddison, A.~Mnih, and Y.~W. Teh.
\newblock The concrete distribution: A continuous relaxation of discrete random
  variables.
\newblock {\em arXiv preprint arXiv:1611.00712}, 2016.

\bibitem{magliacane2018domain}
S.~Magliacane, T.~van Ommen, T.~Claassen, S.~Bongers, P.~Versteeg, and J.~M.
  Mooij.
\newblock Domain adaptation by using causal inference to predict invariant
  conditional distributions.
\newblock In {\em Advances in Neural Information Processing Systems}, pages
  10846--10856, 2018.

\bibitem{murdoch2019definitions}
W.~J. Murdoch, C.~Singh, K.~Kumbier, R.~Abbasi-Asl, and B.~Yu.
\newblock Definitions, methods, and applications in interpretable machine
  learning.
\newblock {\em Proceedings of the National Academy of Sciences},
  116(44):22071--22080, 2019.

\bibitem{sorokina2008detecting}
D.~Sorokina, R.~Caruana, M.~Riedewald, and D.~Fink.
\newblock Detecting statistical interactions with additive groves of trees.
\newblock In {\em Proceedings of the 25th international conference on Machine
  learning}, pages 1000--1007, 2008.

\bibitem{staiger1989cacgtg}
D.~Staiger, H.~Kaulen, and J.~Schell.
\newblock A cacgtg motif of the antirrhinum majus chalcone synthase promoter is
  recognized by an evolutionarily conserved nuclear protein.
\newblock {\em Proceedings of the National Academy of Sciences},
  86(18):6930--6934, 1989.

\bibitem{tsang2020feature}
M.~Tsang, D.~Cheng, H.~Liu, X.~Feng, E.~Zhou, and Y.~Liu.
\newblock Feature interaction interpretability: A case for explaining
  ad-recommendation systems via neural interaction detection.
\newblock In {\em International Conference on Learning Representations}, 2020.

\bibitem{tsang2017detecting}
M.~Tsang, D.~Cheng, and Y.~Liu.
\newblock Detecting statistical interactions from neural network weights.
\newblock {\em arXiv preprint arXiv:1705.04977}, 2017.

\bibitem{velivckovic2017graph}
P.~Veli{\v{c}}kovi{\'c}, G.~Cucurull, A.~Casanova, A.~Romero, P.~Lio, and
  Y.~Bengio.
\newblock Graph attention networks.
\newblock {\em arXiv preprint arXiv:1710.10903}, 2017.

\bibitem{welinder2010caltech}
P.~Welinder, S.~Branson, T.~Mita, C.~Wah, F.~Schroff, S.~Belongie, and
  P.~Perona.
\newblock Caltech-ucsd birds 200.
\newblock 2010.

\bibitem{zheng2018dags}
X.~Zheng, B.~Aragam, P.~K. Ravikumar, and E.~P. Xing.
\newblock Dags with no tears: Continuous optimization for structure learning.
\newblock In {\em Advances in Neural Information Processing Systems}, pages
  9472--9483, 2018.

\bibitem{zheng2019learning}
X.~Zheng, C.~Dan, B.~Aragam, P.~Ravikumar, and E.~P. Xing.
\newblock Learning sparse nonparametric dags.
\newblock {\em arXiv preprint arXiv:1909.13189}, 2019.

\bibitem{zhou2018graph}
J.~Zhou, G.~Cui, Z.~Zhang, C.~Yang, Z.~Liu, L.~Wang, C.~Li, and M.~Sun.
\newblock Graph neural networks: A review of methods and applications.
\newblock {\em arXiv preprint arXiv:1812.08434}, 2018.

\end{thebibliography}

\newboolean{add_appendix}
\setboolean{add_appendix}{true}
\ifthenelse{\boolean{add_appendix}} {
\newpage
\appendix
\section{Appendix}
\paragraph{Additional \methodName Details }: We want to point out that modeling a binary graph with prior matrix $\gamma$ allows us to have a probabilistic interpretation of each edge in the graphs we learn. Additionally, the combination of a global (population level) dependency graph along with sample level attention (on each edge learnt by GAT) allows us to model the global binary dependency  structure, then refined with local (sample level) fine-tuning regarding which edges were most important for a specific prediction.
An easy analogy: learning without a structure is like a tourist without a map. The graph we learn is the map for everyone and how to use the map depends on each traveler’s choice (being modelled as graph attention for each sample).

\subsection{Simulated 2D Classification Datasets}
We first evaluate our model on a simple 2D dataset. We consider the binary classification of the dataset shown in Figure~\ref{fig:2d_graphs}(a). The task is to classify Black Points (denoted as Class A) from the Green Points (Class B). We generate Gaussian dataset samples from Class A using $\bX_{A}\sim\mathcal{N}(0,\boldsymbol{\Omega}_{A}^{-1})$ and $\bX_{B}\sim\mathcal{N}(0,\boldsymbol{\Omega}_{B}^{-1})$.  Here, $\boldsymbol{\Omega}_{A}=
(\begin{smallmatrix}
1.0 & 0.99\\
0.99 & 1.0
\end{smallmatrix})
$ and 
$\boldsymbol{\Omega}_{B}=
(\begin{smallmatrix}
1.0 & -0.99\\
-0.99 & 1.0
\end{smallmatrix})
$. 
In this case, both $\boldsymbol{x}_1$ and $\boldsymbol{x}_2$ have identical marginal distributions. The model needs to incorporate the interactions between $\boldsymbol{x}_1$ and $\boldsymbol{x}_2$ i.e. both the input variables to classify correctly. Note that this data has two types of interactions, in Class A the two features interact positively, while the points from Class B interact negatively. As shown in  Figure~\ref{fig:2d_graphs}, we are successfully able to discover the correct graphs. We compare the class specific graphs from M-{\methodName} in Figure ~\ref{fig:2d_graphs}(c)(Class A) and Figure ~\ref{fig:2d_graphs}(d)(Class B) learnt using the {\methodName} in Figure ~\ref{fig:2d_graphs}(b).  If we use a single shared graph, we are unable to recover a true representative set of interactions. Instead, our M-{\methodName} variation allows us to incorporate this class specific variability.  In terms of classification performance, {\methodName} is able to achieve  a classification AUC of $0.9956$, M-{\methodName} of $0.9956$ and a 2 layer MLP is able to achieve $0.9957$. 

\paragraph{Hyper parameters:} We use  hidden size $task_H=8$, layers $task_L=2$, heads $K=2$, structure learner's hidden size $func_H=8$ and $d_{pos}=4$. We compare to an MLP with $2$ hidden layers and hidden size $16$. We use $\ell_{sparse}=0.0$ and $\ell_{struct}=10.0$. We train the models for $250$ epochs. %
To optimize the model, we use Adam optimizer with learning rate $0.001$. We report the average performance for $3$ random seeds.

\begin{figure*}[hb]
\begin{center}
\includegraphics[width=0.8\textwidth]{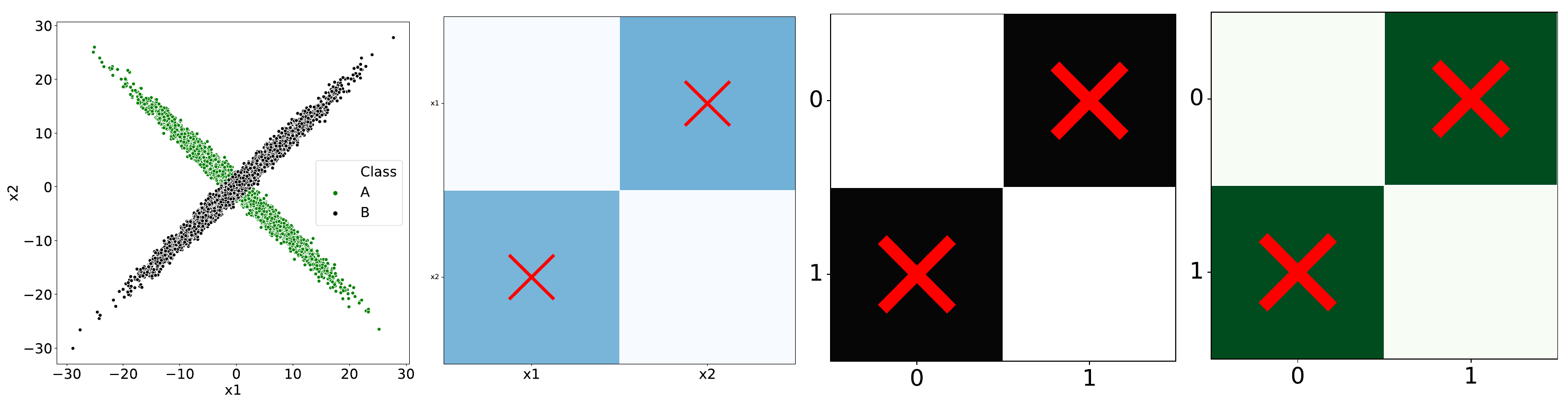}
\end{center}
\vspace{-0.1in}
\caption{On Simulation: Learnt Graphs using \methodName: (a) 2D Gaussian Data (b)  \methodName; M-\methodName (c) Graph from Class A and (d) Graph from Class B. \fix{this needs to be re-done.. and needs a much more comprehensive analysis on the dataset, I think we need to investigate carefully on all possible cases ( 0, 0.5, 1) via this type of 2D example to understand the method better. In addition,we should use simulation data to understand zij = 0.5 cases.  please standardize the names of models }  
\label{fig:2d_graphs}
}
\end{figure*}

\subsection{Simulated $p$-D Classification Datasets}
In this section, we present more visualization and results for the $p$-dimensional Gaussian Datasets. In Table ~\ref{tab:shared_sim_v2} we provide detailed results, reporting the average and standard deviation across $5$ random seeds for classification.  Table~\ref{tab:shared_sim_graph_v2} reports the average and standard deviation for graph AUC across $5$ random seeds.

{\bf Visualization:}Figure~\ref{fig:simulation_10} shows the different graphs learnt by ablations and variations of \methodName. In Figure~\ref{fig:simulation_10}(b), We show that the learnt graph for the single graph case \methodName corresponds to the shared graph entries in the ground truth graph. Figure~\ref{fig:simulation_10}(f) indicates the importance of the self-supervision loss as without this loss even with the sparsity assumption, we are unable to recover any correct edges at all.  Similarly, Figure~\ref{fig:sim_20} shows the discovered graphs for multi graph variation M-\methodName for $p=20$.

\begin{table*}[tbh]

   \centering
    \begin{tabular}{|l|c|c|c|c|}\hline

           \makecell{Models, Baselines, \\  Ablations and Variations} & \makecell{AUC \\ ($p=5$)} & \makecell{AUC \\ ($p=10$)} & \makecell{AUC \\ ($p=20$)}  \\\hline\hline
        {\methodName} & 0.7132($\pm$0.006)  & 0.7145($\pm$0.026) &  \textbf{0.8490}($\pm$ 0.019)  \\\hline
        {\methodNameNSparse}  & 0.7139($\pm$0.006) & 0.7162($\pm$0.029) &  0.8479($\pm$ 0.019) \\\hline
        {\methodNameNS}  & 0.7144($\pm$0.006) & \textbf{0.719}($\pm$ 0.029) & 0.8471($\pm$ 0.020) \\\hline
        {\methodNameGCN} &  0.6324($\pm$ 0.028) & 0.6191($\pm$ 0.030) & 0.6635($\pm$ 0.032)\\\hline
        M-{\methodName} & \textbf{0.7147}($\pm$ 0.005) & 0.7142($\pm$ 0.033)  & 0.8467($\pm$0.018) \\\hline
        M-{\methodNameNSparse} &  0.7135($\pm$ 0.006) & 0.7153($\pm 0.029$)  & 0.8443($\pm$0.018)  \\\hline
        M-{\methodNameNS} &  0.7139($\pm$ 0.005) & 0.7165($\pm$0.029) & 0.8453($\pm$0.020) \\\hline
 M-{\methodNameGCN} & 0.6479($\pm$0.012)  & 0.6141($\pm$0.038) & 0.6614($\pm$0.071)  \\\hline\hline

          GAT-FC & 0.7137($\pm$0.006) & 0.7051($\pm$ 0.014) & 0.8489($\pm$ 0.020)  \\\hline
         MLP & 0.7161($\pm$ 0.006) & 0.7193($\pm$0.029) & \textbf{0.8548}($\pm$0.017)  \\\hline
        QDA & \textbf{0.7178}($\pm$0.004) & \textbf{0.7252}($\pm$0.026) & 0.8215($\pm$ 0.021)  \\\hline 
    \end{tabular}
    \caption{Classification Results Area under Curve (AUC) on test data and Evaluation on Graph Estimations  for simulation datasets averaged across $5$ random seeds. %
    }
    \label{tab:shared_sim_v2}
\end{table*}

 \begin{table*}[tbh]
   \centering
    \begin{tabular}{|l|l|l|l|}\hline

           \makecell{Models, Baselines, \\  Ablations and Variations}  & \makecell{Graph-AUC \\ ($p=5$)} & \makecell{Graph-AUC \\ ($p=10$)} & \makecell{Graph-AUC \\ ($p=20$)} \\\hline\hline
        {\methodName}  & \textbf{1.0}($\pm$0.000) & 0.9979($\pm$0.004) & 0.9957($\pm$0.004) \\\hline
        {\methodNameNSparse}   &  0.4667($\pm$0.155)
 & 0.4713($\pm$0.164) &  0.4462($\pm$ 0.095)\\\hline
        {\methodNameNS}   & 0.4521 ($\pm$0.160) & 0.4602($\pm$
0.074) &  0.5297($\pm$0.047)
        \\\hline
        {\methodNameGCN}  & \textbf{1.0}($\pm$0.000) & 0.9997($\pm$0.0014) & 0.9973($\pm$0.0053) \\\hline
        M-{\methodName}  & \textbf{1.0}($\pm$0.000)  & \textbf{1.0}($\pm$0.000) & \textbf{1.0}($\pm$0.000) \\\hline
        M-{\methodNameNSparse}  & 0.3396($\pm$0.228)
& 0.5680($\pm$0.119)
  & 0.5912($\pm$0.080) \\\hline
        M-{\methodNameNS}  & 0.6021($\pm$0.177) & 0.4567($\pm$ 0.059) & 0.467($\pm$0.065)\\\hline
 M-{\methodNameGCN}   & \textbf{1.0}($\pm$0.000) & \textbf{1.0}($\pm$0.000) & \textbf{1.0}($\pm$0.000) \\\hline
  
        NID  & 0.6250($\pm$0.088) & 0.6526($\pm$ 0.026) & 0.6300($\pm$0.026) \\\hline

    \end{tabular}
    \caption{Evaluation on Graph Estimations  for simulation datasets averaged across $5$ random seeds. %
    }
    \label{tab:shared_sim_graph_v2}
\end{table*}

\begin{figure*}[tbh]
\begin{center}
\setlength\tabcolsep{1.5pt} 
\begin{tabular}{cccc}
\includegraphics[trim=0 10 0 190, clip,scale=0.24]{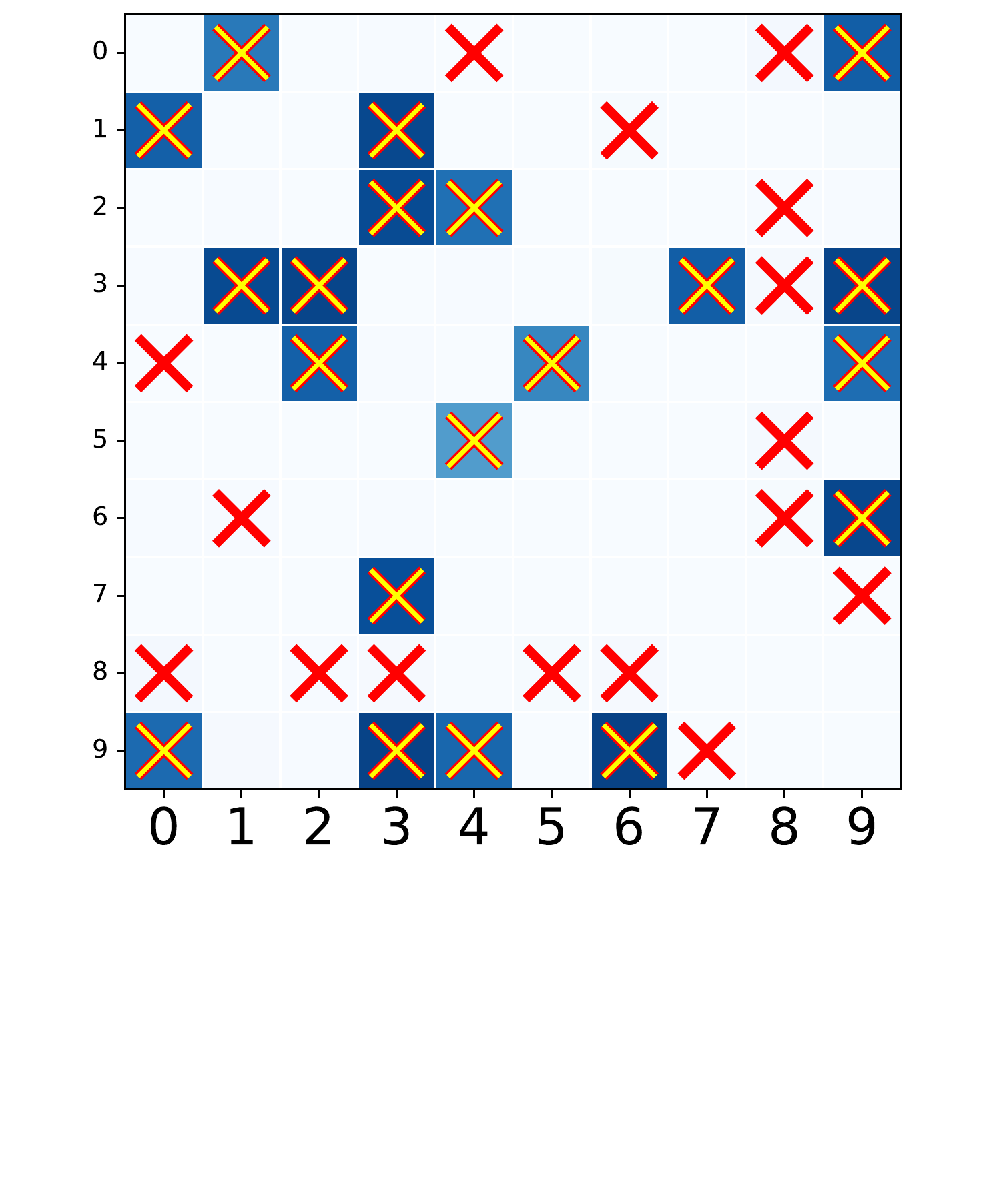}&
\includegraphics[trim=0 10 0 190, clip,scale=0.24]{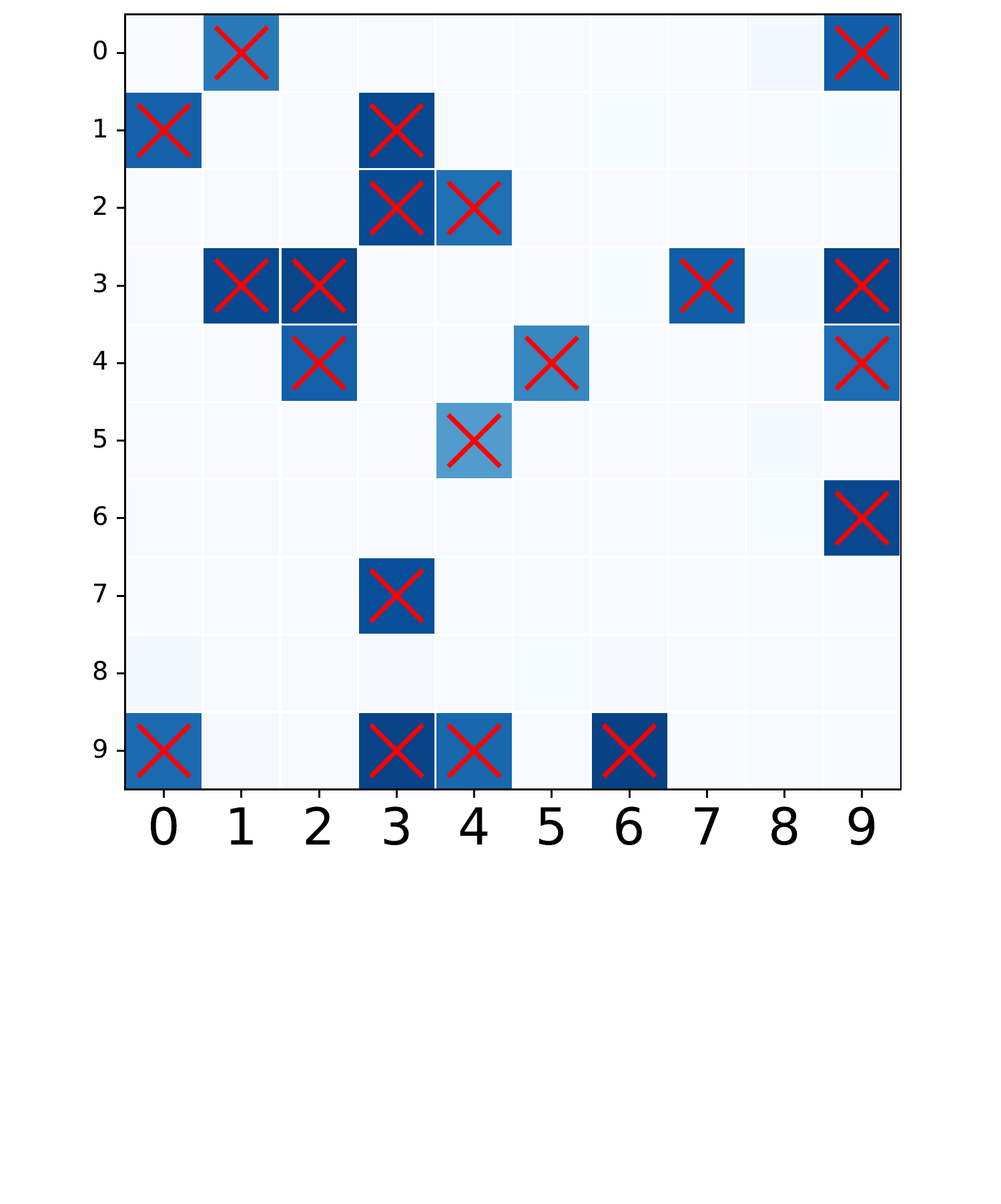}&
\includegraphics[trim=0 10 0 190, clip,scale=0.24]{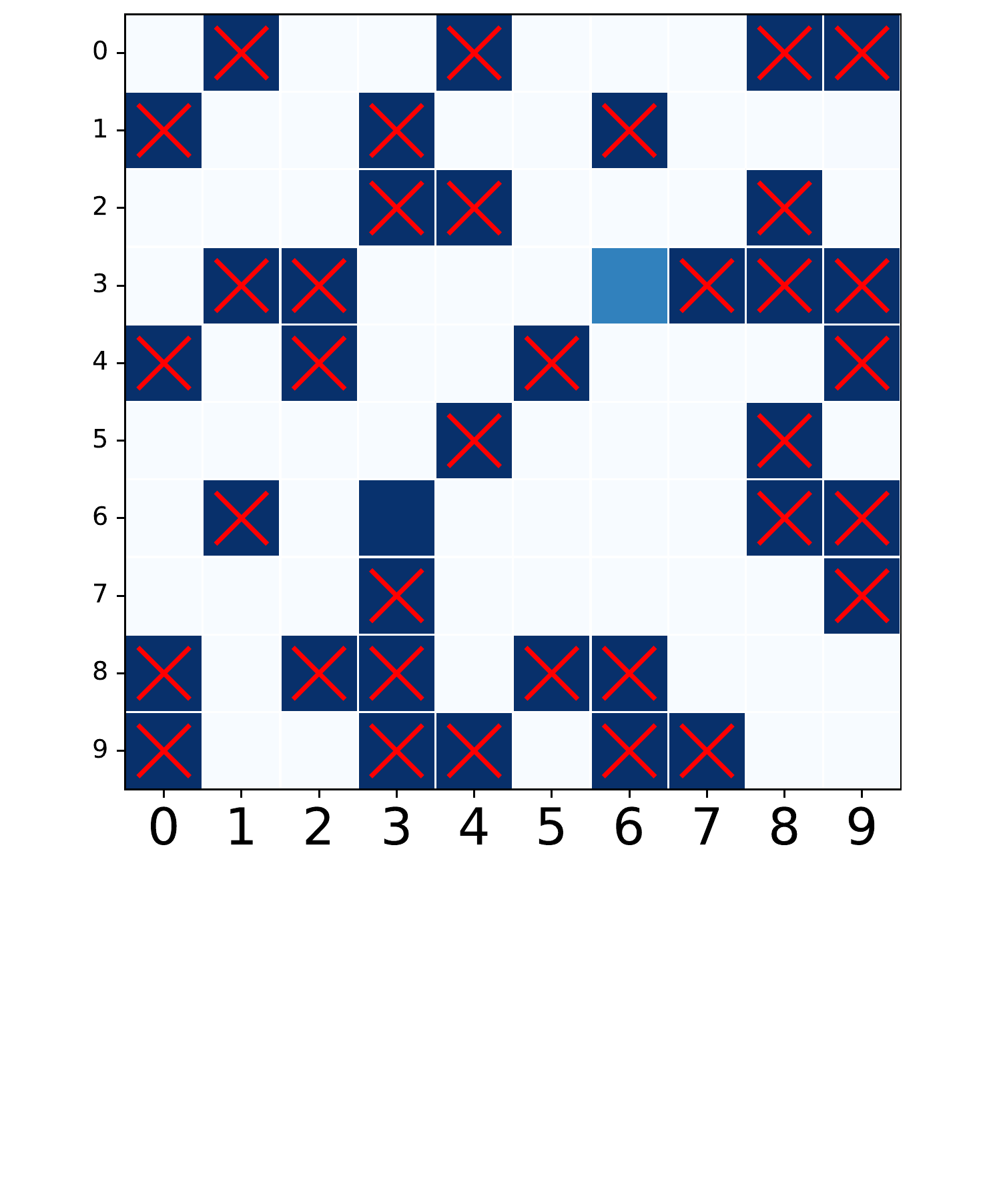}&
\includegraphics[trim=0 10 0 190,clip,scale=0.24]{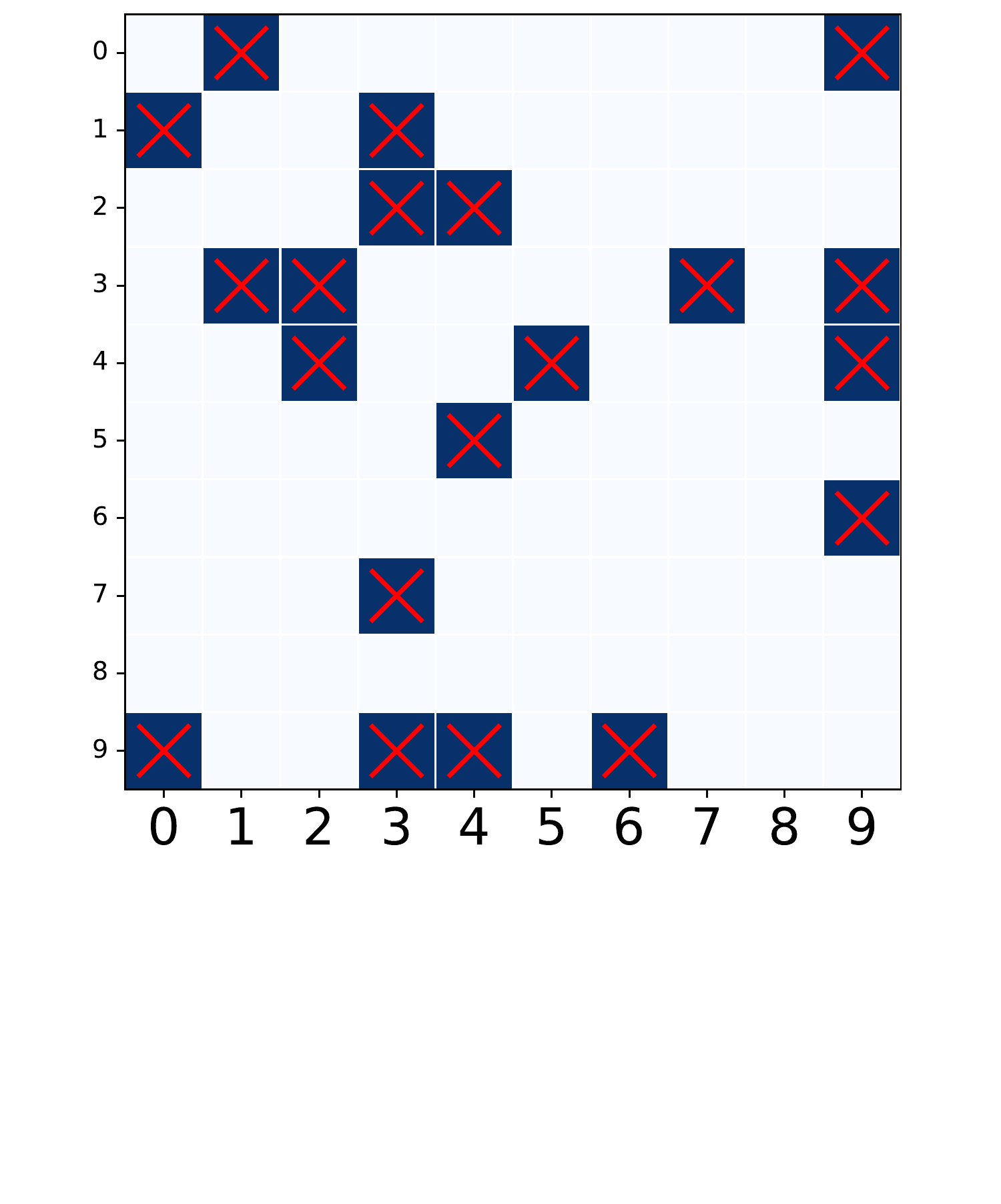}\vspace{-12mm}\\
(a) & (b) & (c)  & (d)\\
\includegraphics[trim=0 10 0 190, clip,scale=0.24]{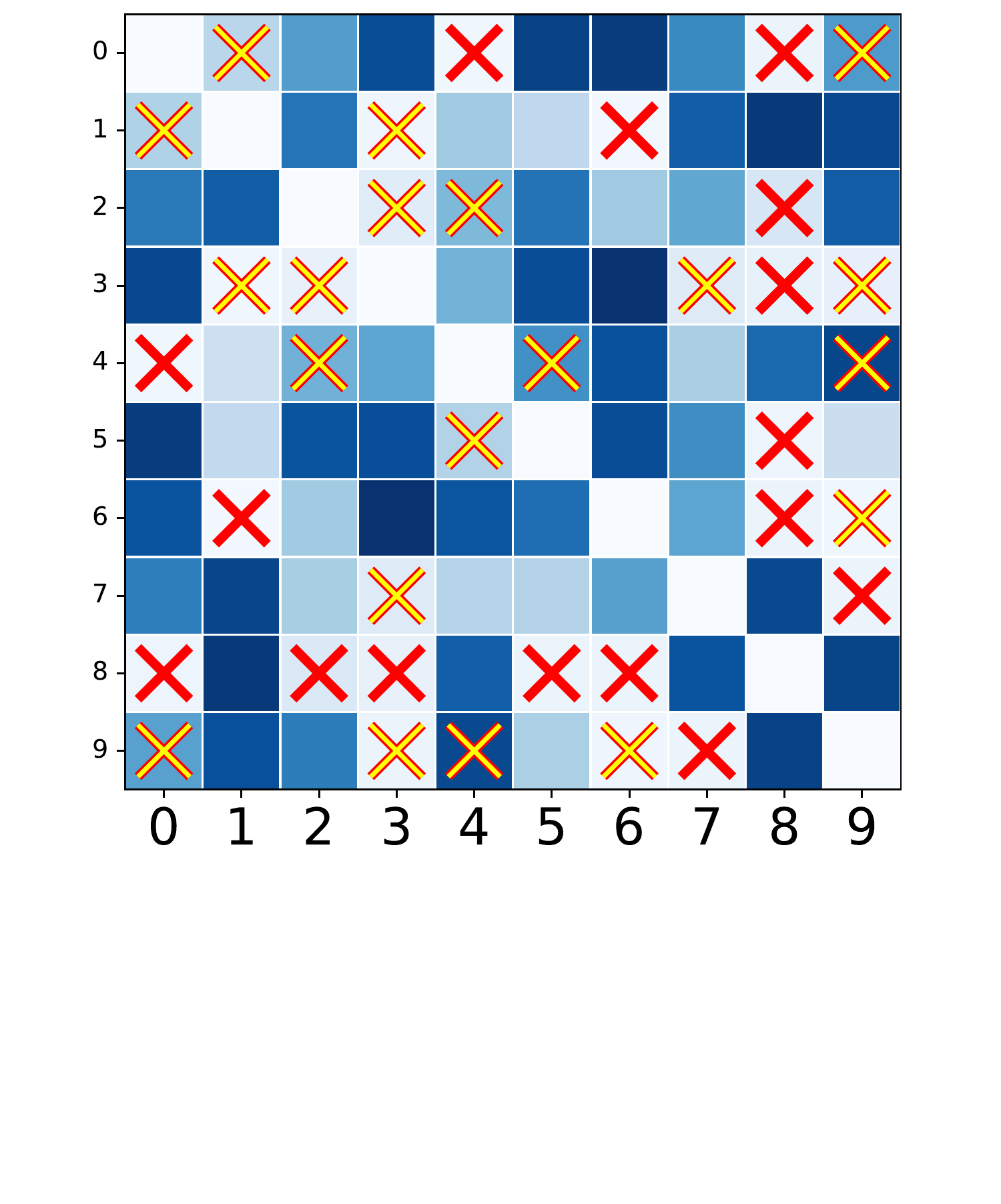}&
\includegraphics[trim=0 10 0 190, clip,scale=0.24]{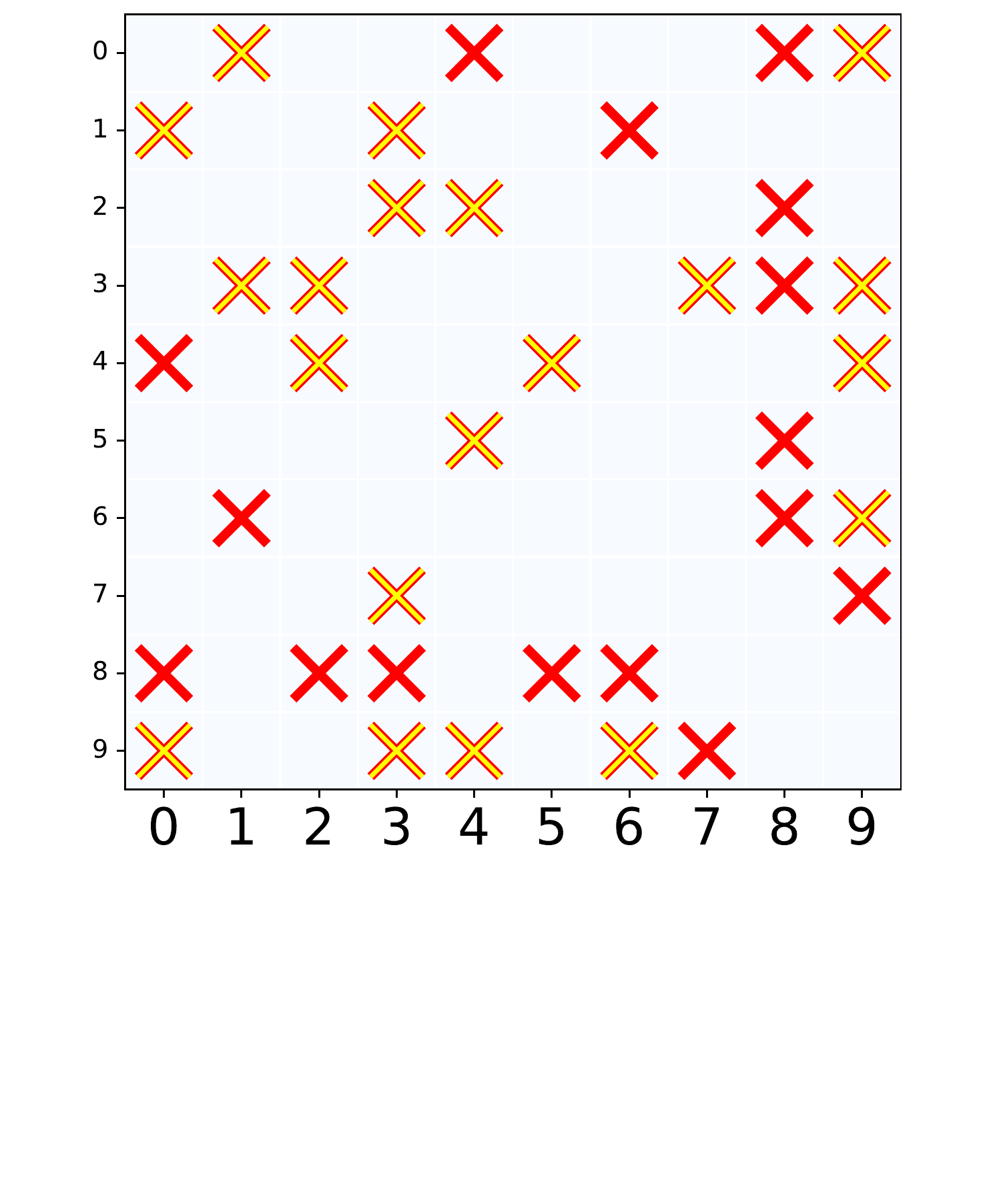}&
\includegraphics[trim=0 10 0 190, clip,scale=0.24]{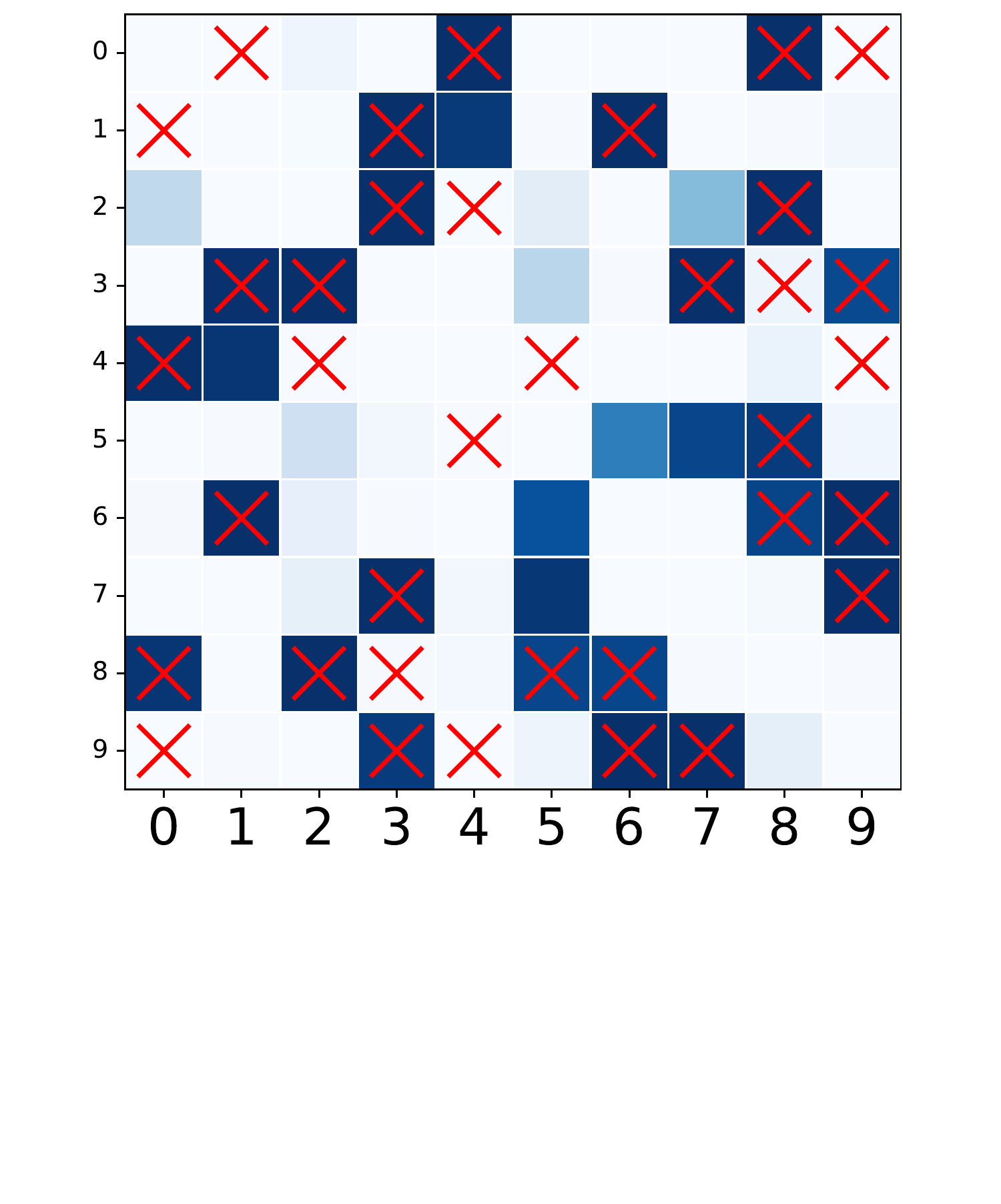}&
\includegraphics[trim=0 10 0 190, clip,scale=0.24]{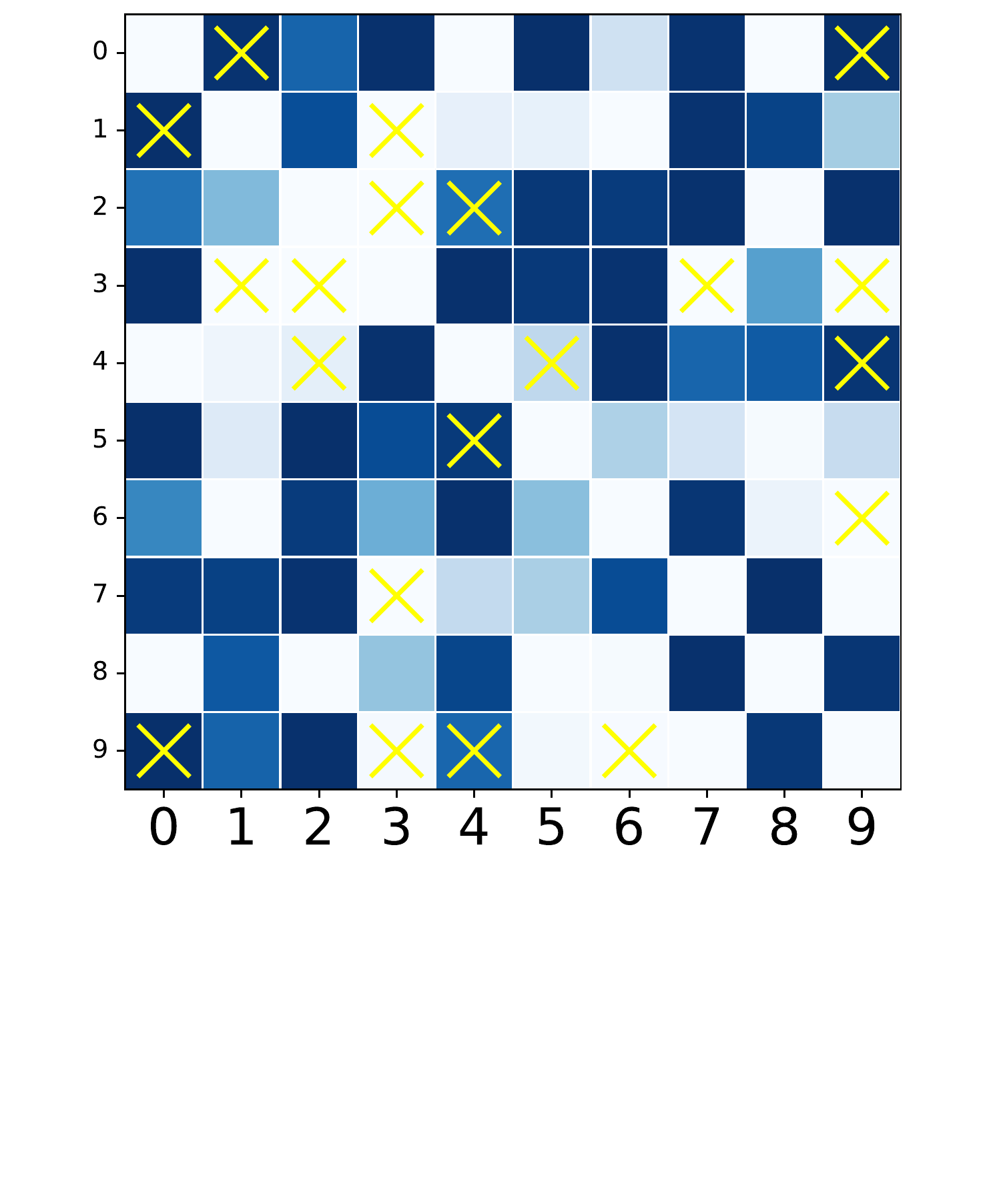}\vspace{-13mm}\\
(e) & (f) & (g) & (h) \\
\end{tabular}
\end{center}
\caption{On Simulation: 
Heat maps of pairwise interaction strengths learnt by {\methodName} framework on  for simulation datasets $p=10$. Cross-marks indicate ground truth interactions. Red cross marks indicate Class $A$ and Yellow cross marks indicate Class $B$. (a), Graph learnt by {\methodName} (b) Graph learnt by {\methodName} in comparison to the graph edges common to both classes. Here, we treat the common edges as ground truth indicated by red cross marks. (c) and (d) Graph learnt by M-{\methodName}. (c) shows the graph learnt for Class $A$ and (d) shows the graph learnt for Class $B$.  In the second row, (e) shows the graphs learnt by {\methodNameNSparse}, (f) shows the graphs learnt by {\methodNameNS} and $\lambda_{sparse}=0.005$, (g) and (h) show the graphs learnt by M-{\methodNameNSparse}: (h) graph corresponding to Class $A$ and (h) graph corresponding to Class $B$. \label{fig:simulation_10}
}
\end{figure*}

\begin{figure*}[th]
\begin{center}
\setlength\tabcolsep{0.1pt} 
\begin{tabular}{cc}
\includegraphics[trim=0 10 0 190, clip,scale=0.4]{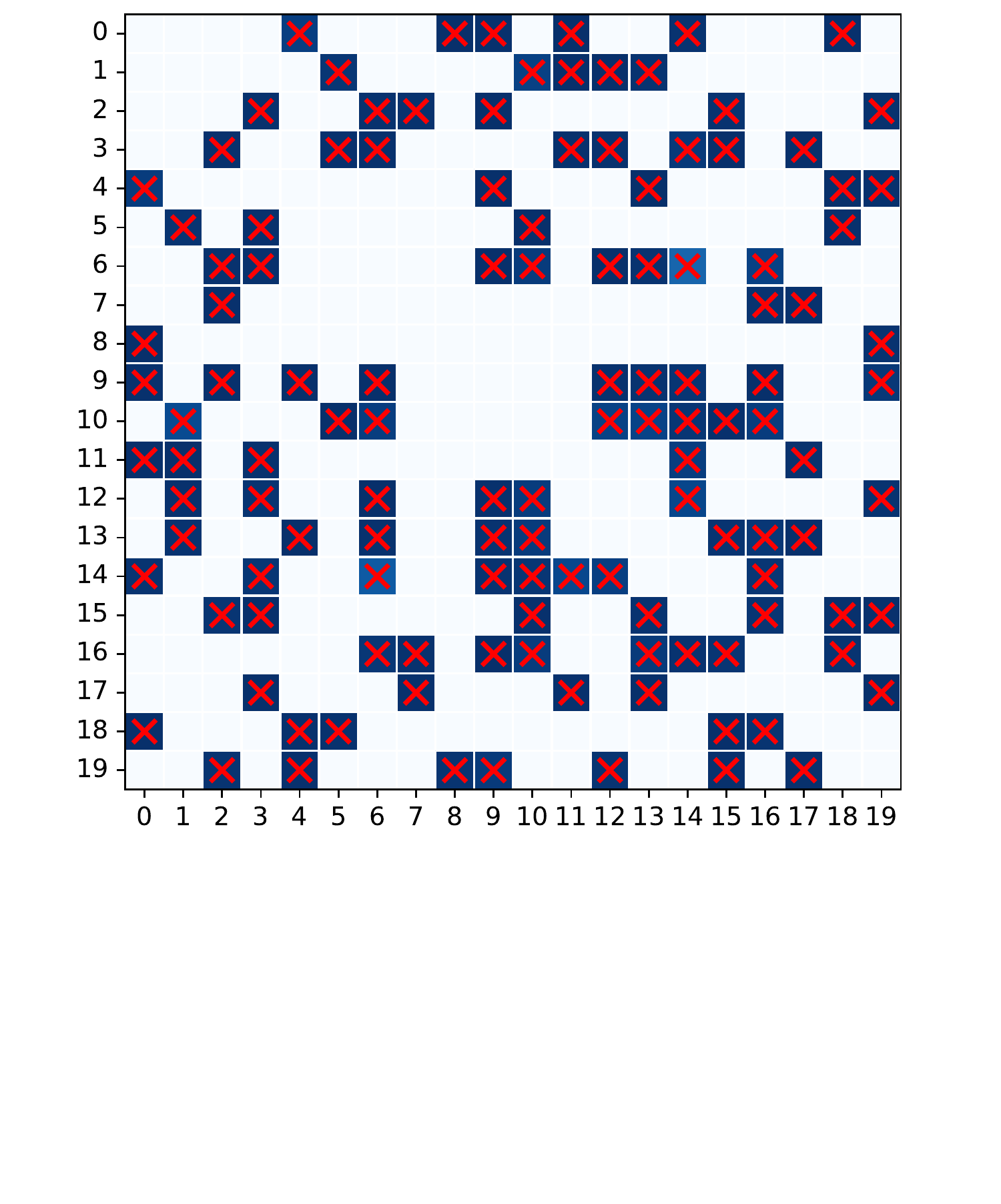} &
\includegraphics[trim=0 10 0 190, clip,scale=0.4]{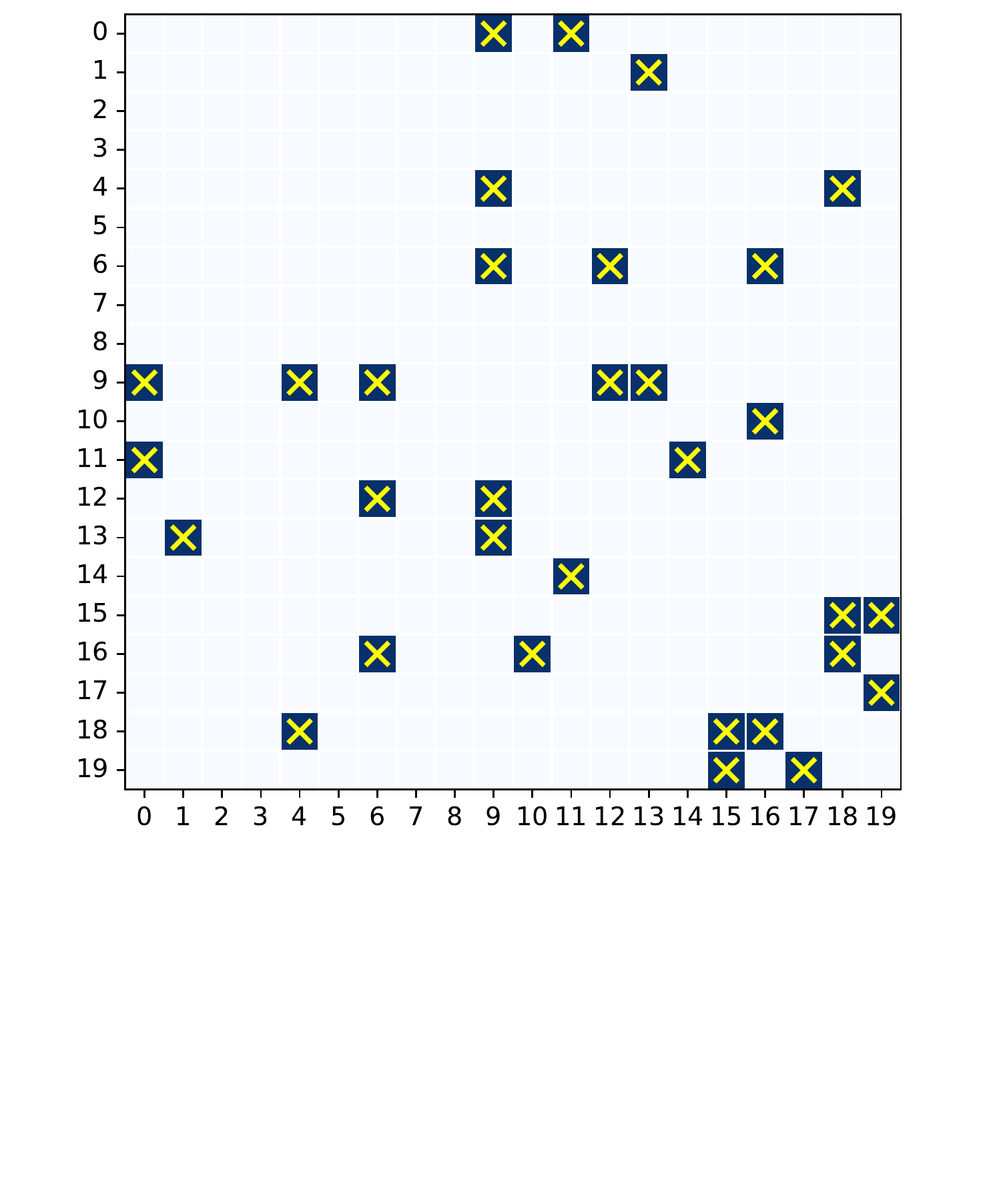} 
\end{tabular}
\end{center}
\vspace{-25mm}
\caption{On Simulation: Learnt graphs for $p=20$, (LEFT) Heat Map for Graph learnt for Class A by M-{\methodName}, Red Cross Marks show the ground truth graphs, and (RIGHT) Heat Map for  Graphs learnt by Class B by M-{\methodName}, Yellow Cross Marks show the ground truth graphs. 
\label{fig:sim_20}
}
\end{figure*}
\subsection{Real World Data: DAG Datasets}
\label{subsec:dag_more}
\paragraph{Additional DAG Visualization}

\begin{figure*}[tbh]
\begin{tabular}{cccc}
  \includegraphics[width=0.23\textwidth]{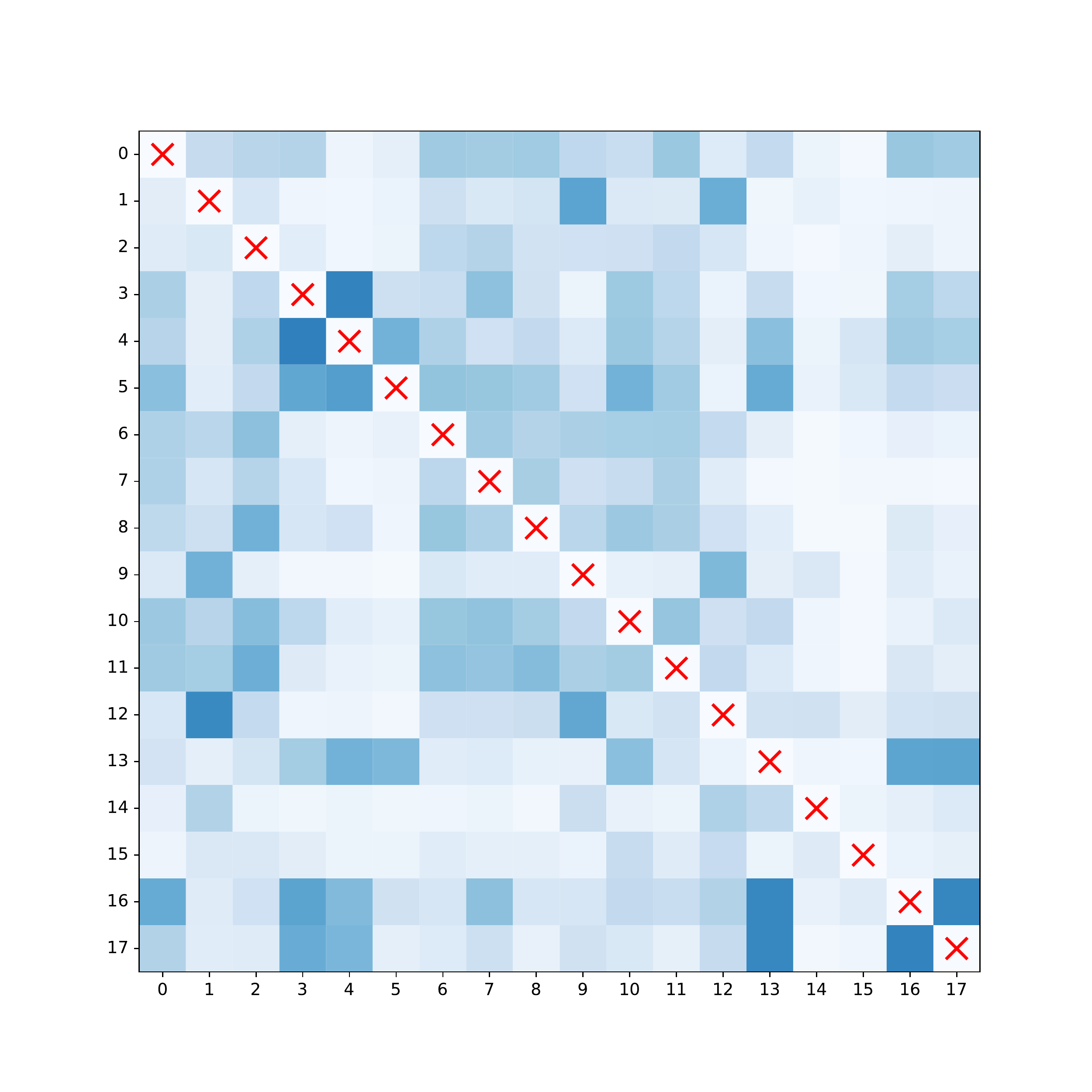} &   \includegraphics[width=0.23\textwidth]{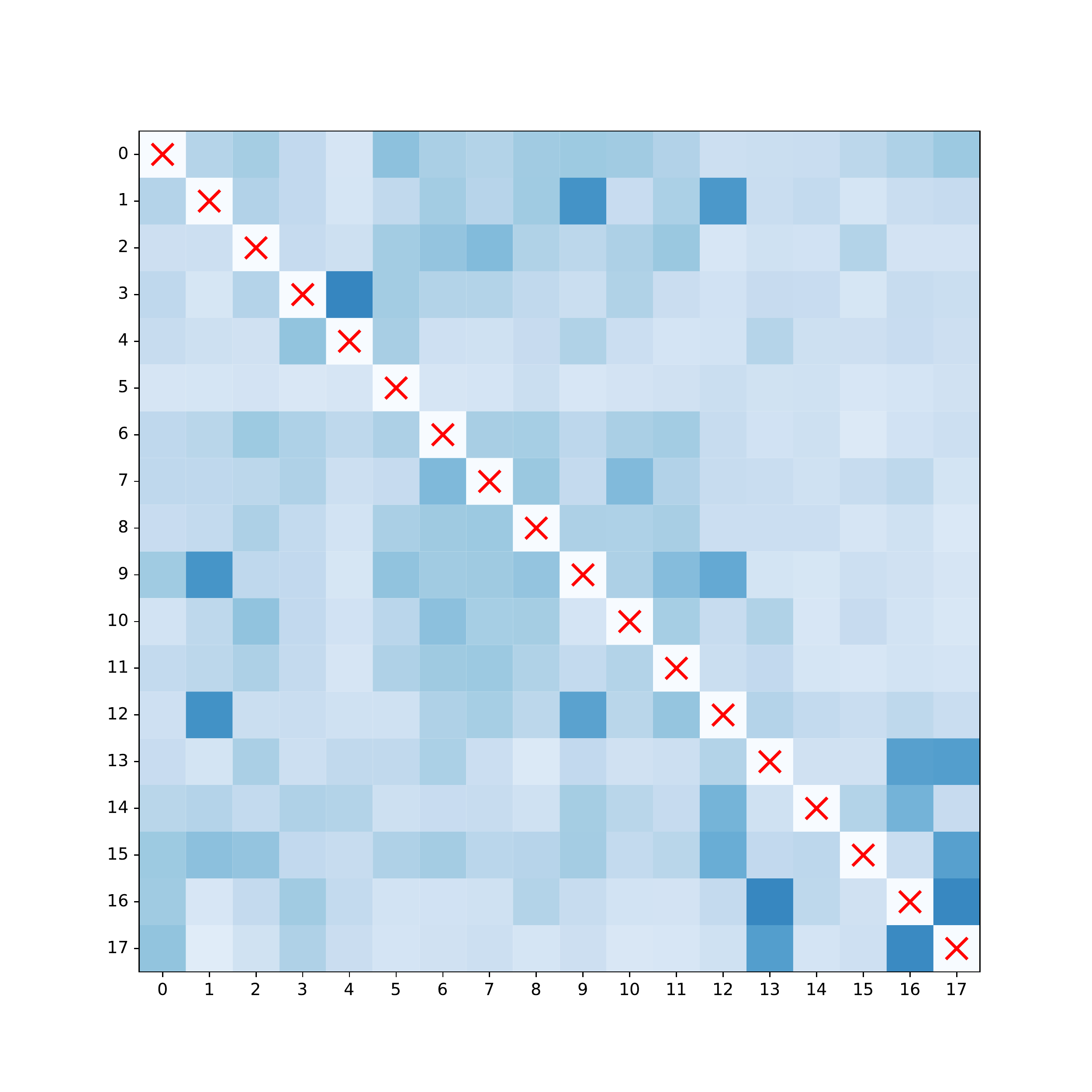} &

 \includegraphics[width=0.23\textwidth]{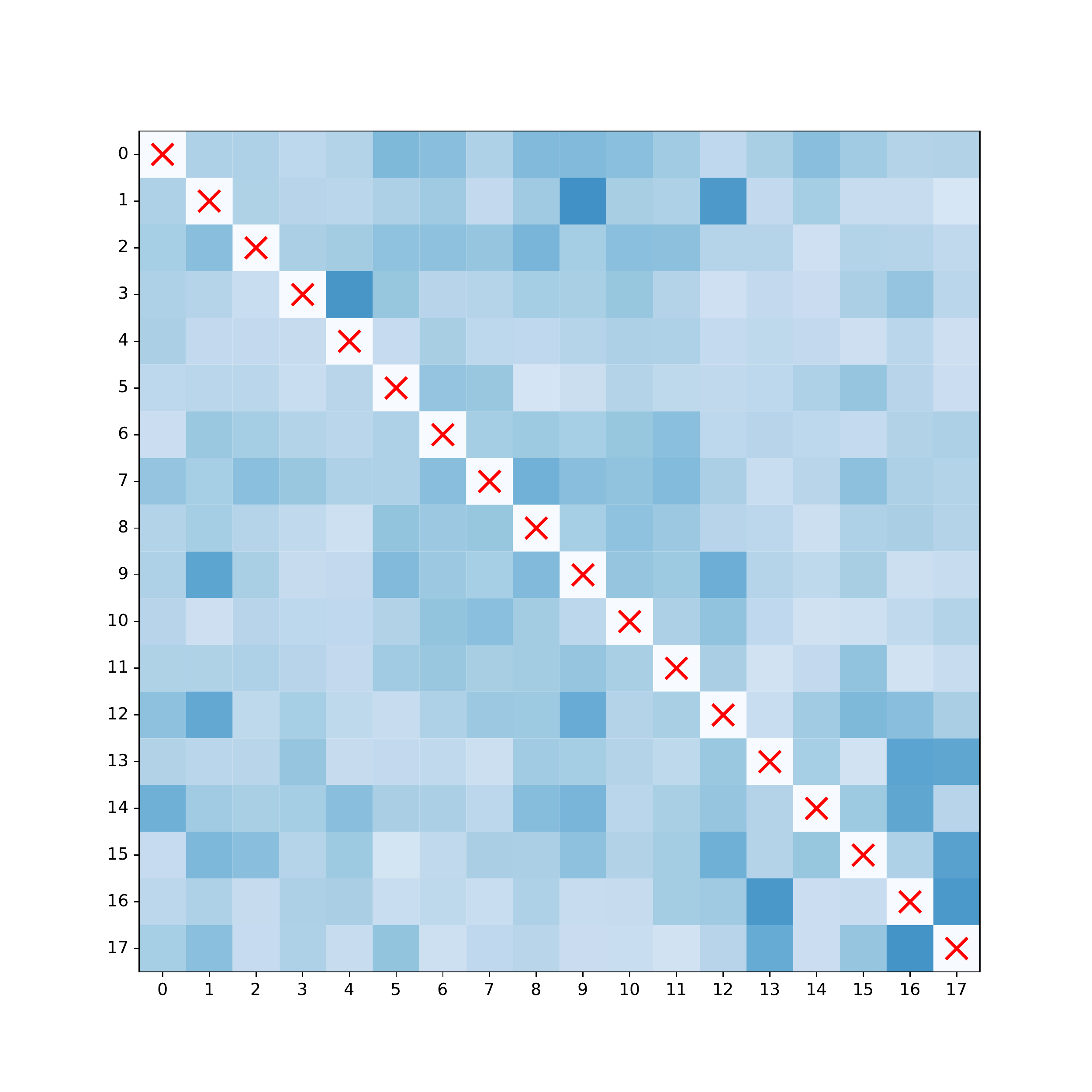} &   \includegraphics[width=0.23\textwidth]{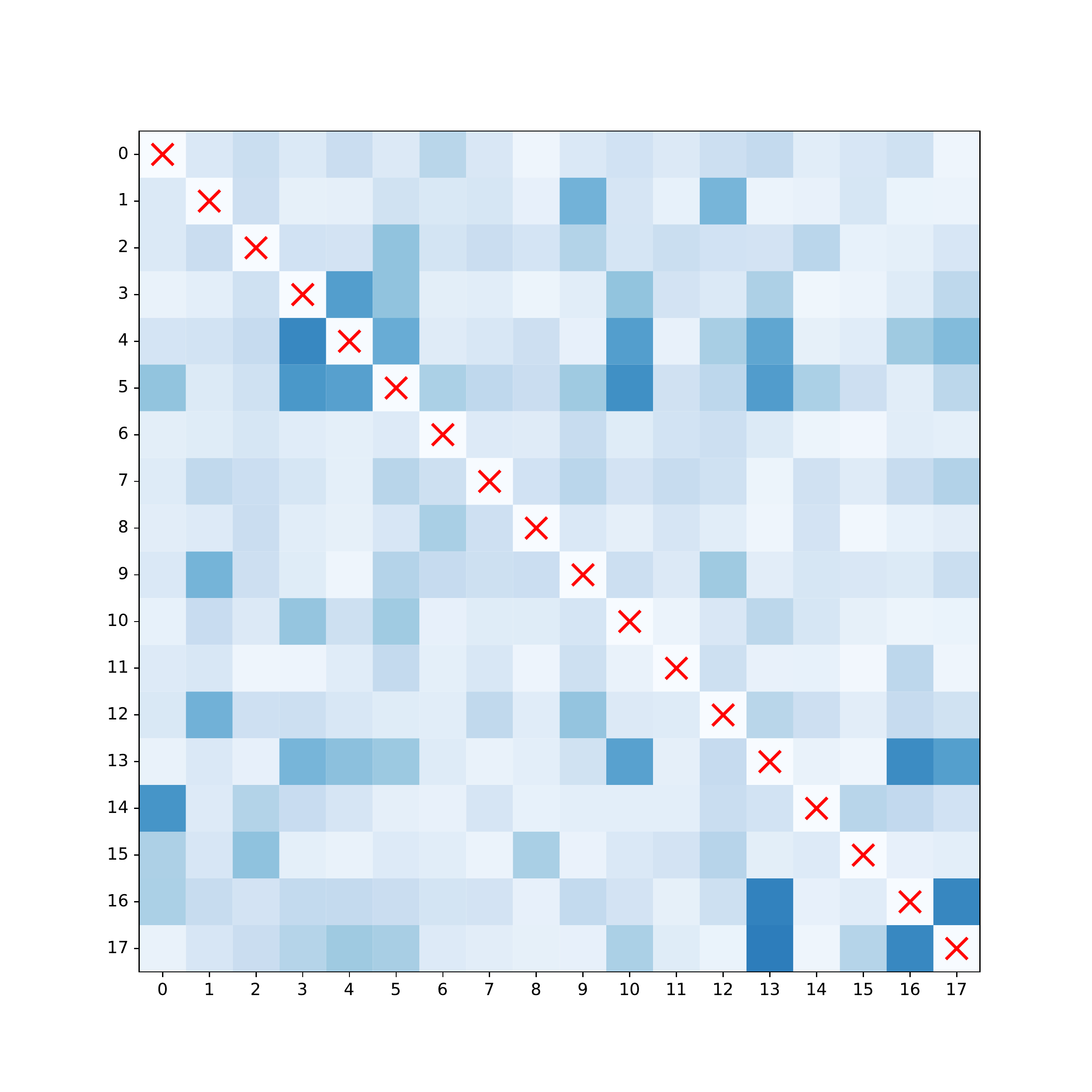} \\
(a) Class 0 & (b) Class 1 & (c) Class 2 & (d) Class 3 \\[6pt]

\end{tabular}

\caption{Visualization of learnt graph by \methodName-L on the Vehicle Dataset. The attributes are compactness, circularity, radius ratio, pr.axis aspect ratio, max length aspect ratio, scatter ratio, elongatedness, pr. axis rectangularity, max length rectangularity, scaled variance along major axis , scaled variance along minor axis, scaled radius of gyration, skewness about minor axis, kurtosis about minor axis and hollows ratio. 
\label{fig:vehicle_l_app}
}
\end{figure*}

\begin{figure*}[tbh]
\begin{tabular}{cccc}
  \includegraphics[width=0.25\textwidth]{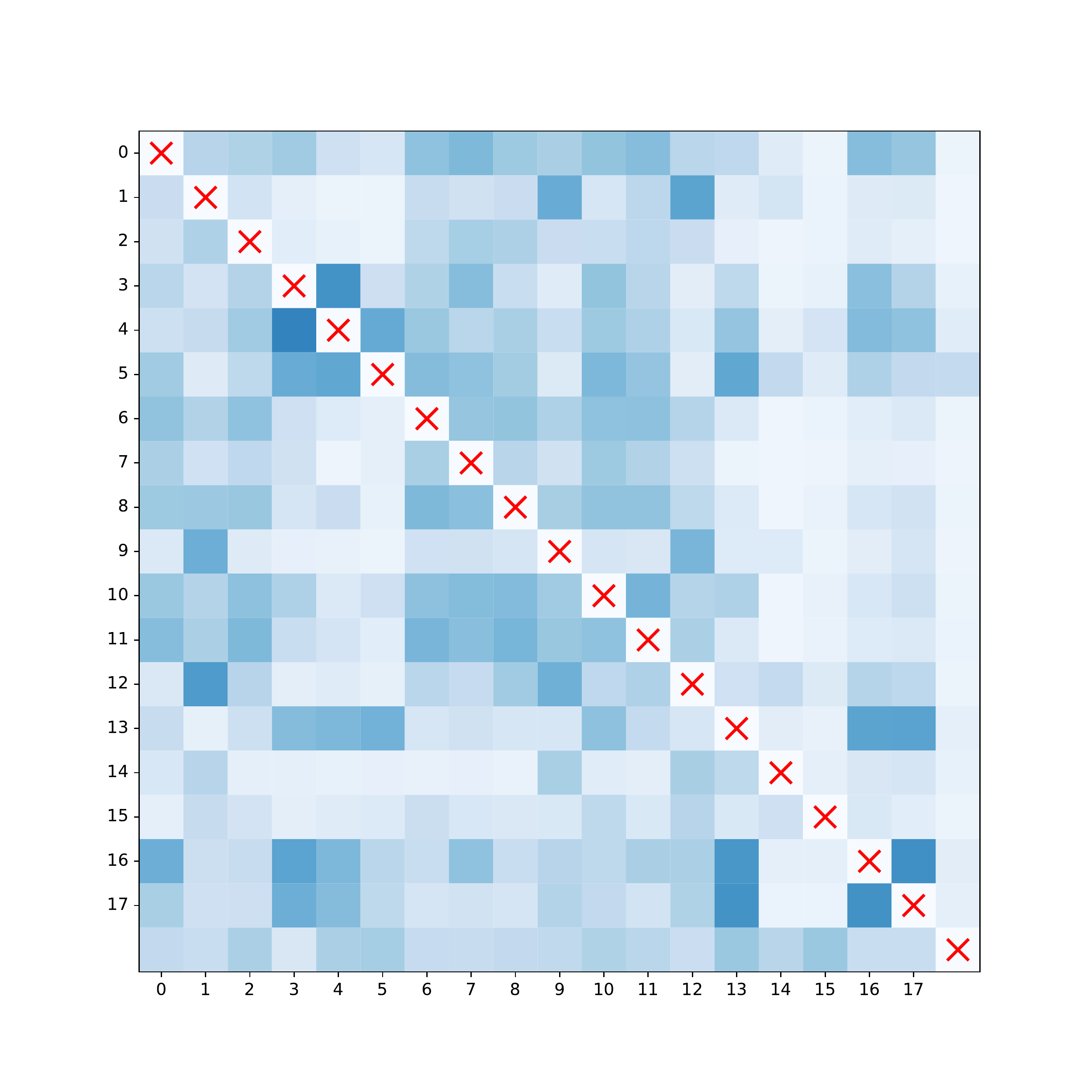} &   \includegraphics[width=0.25\textwidth]{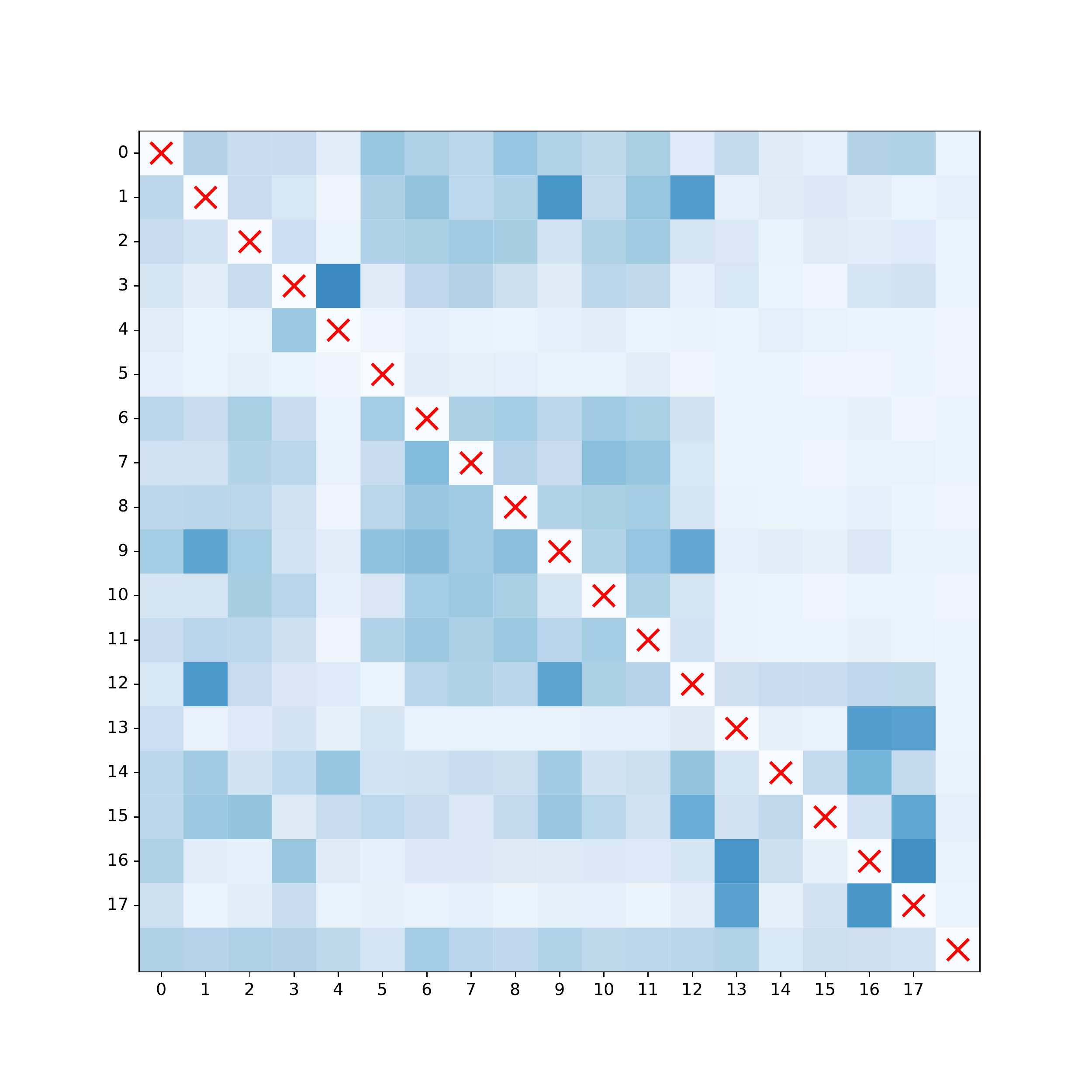} &

 \includegraphics[width=0.25\textwidth]{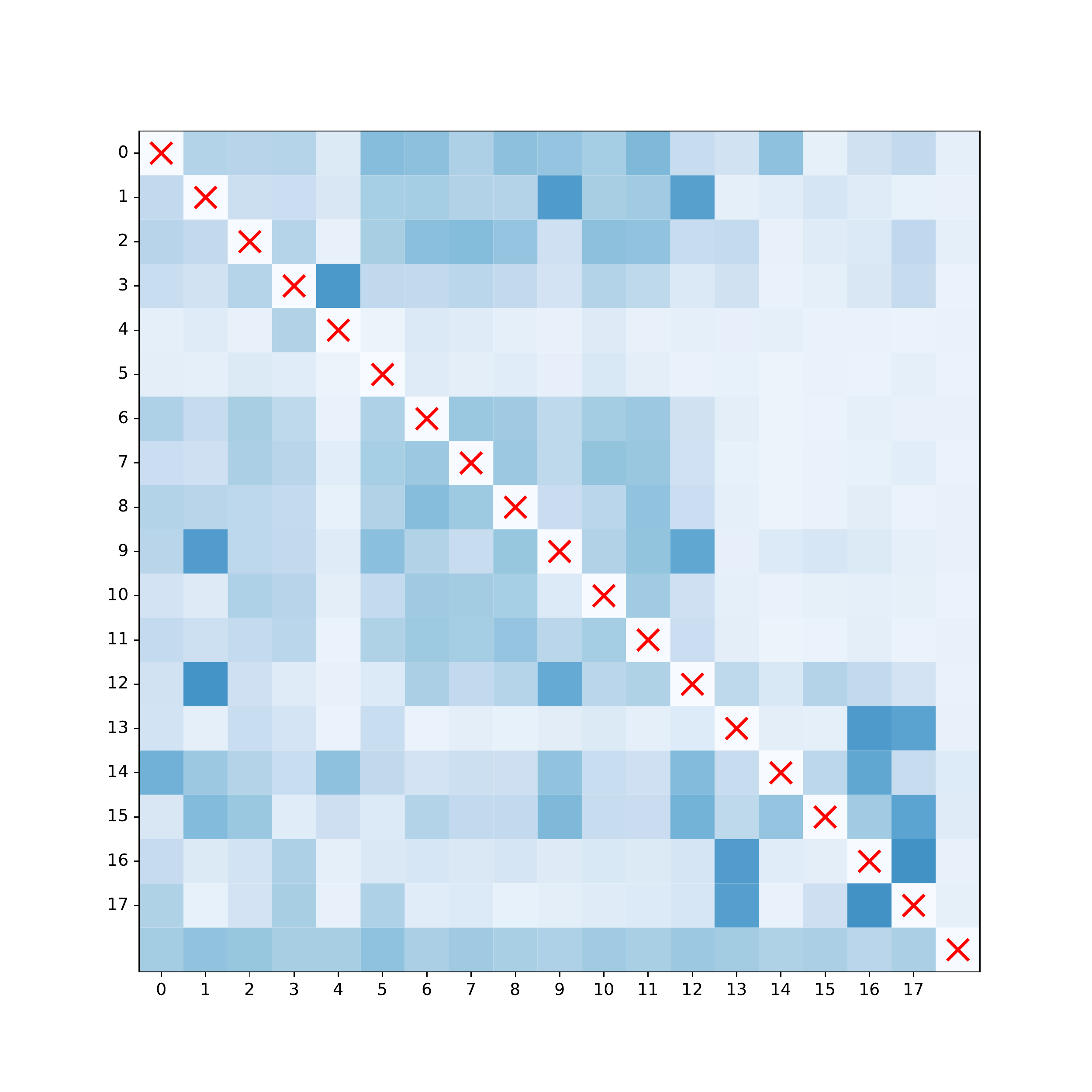} &   \includegraphics[width=0.25\textwidth]{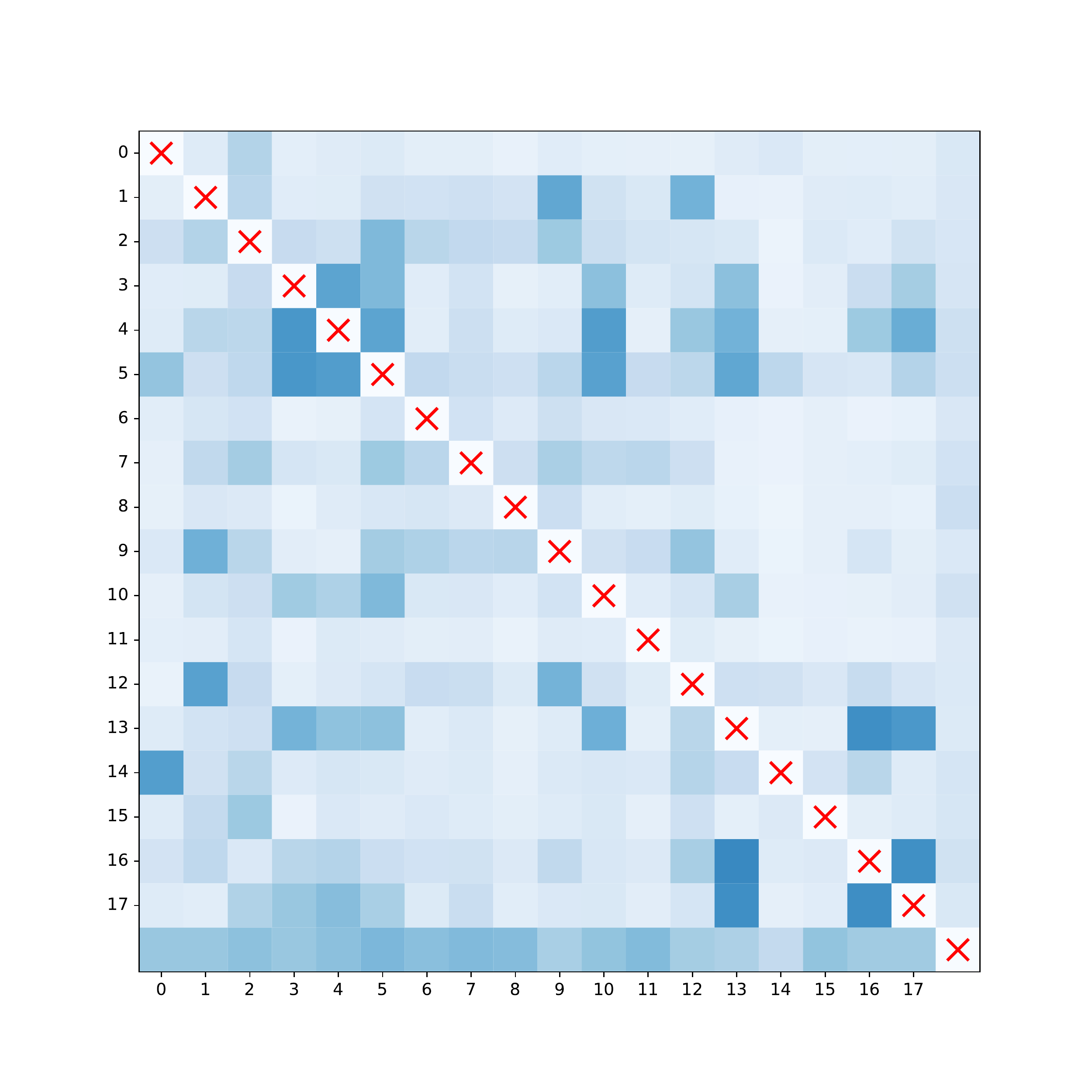} \\
(a) Class 0 & (b) Class 1  & (c) Class 2 & (d) Class 3 \\[6pt]

\end{tabular}

\caption{Visualization of learnt graph by \methodName-L-Y on the Vehicle Dataset
\label{fig:vehicle_l_y_app}
}

\end{figure*}
We include visualizations of the learnt graphs by \methodName and \methodName variations. Figure~\ref{fig:vehicle_l_app} and Figure~\ref{fig:vehicle_l_y_app} show the learnt graphs by Vehicle dataset. The attributes are compactness, circluarity, radius ratio, pr.axis aspect ratio, max length aspect ratio, scatter ratio, elongatedness, pr. axis rectangularity, max length rectangularity, scaled variance along major axis , scaled variance along minor axis, scaled radius of gyration, skewness about minor axis, kurtosis about minor axis and hollows ratio. 

\begin{figure*}[htb]
\begin{center}
\includegraphics[scale=0.50]{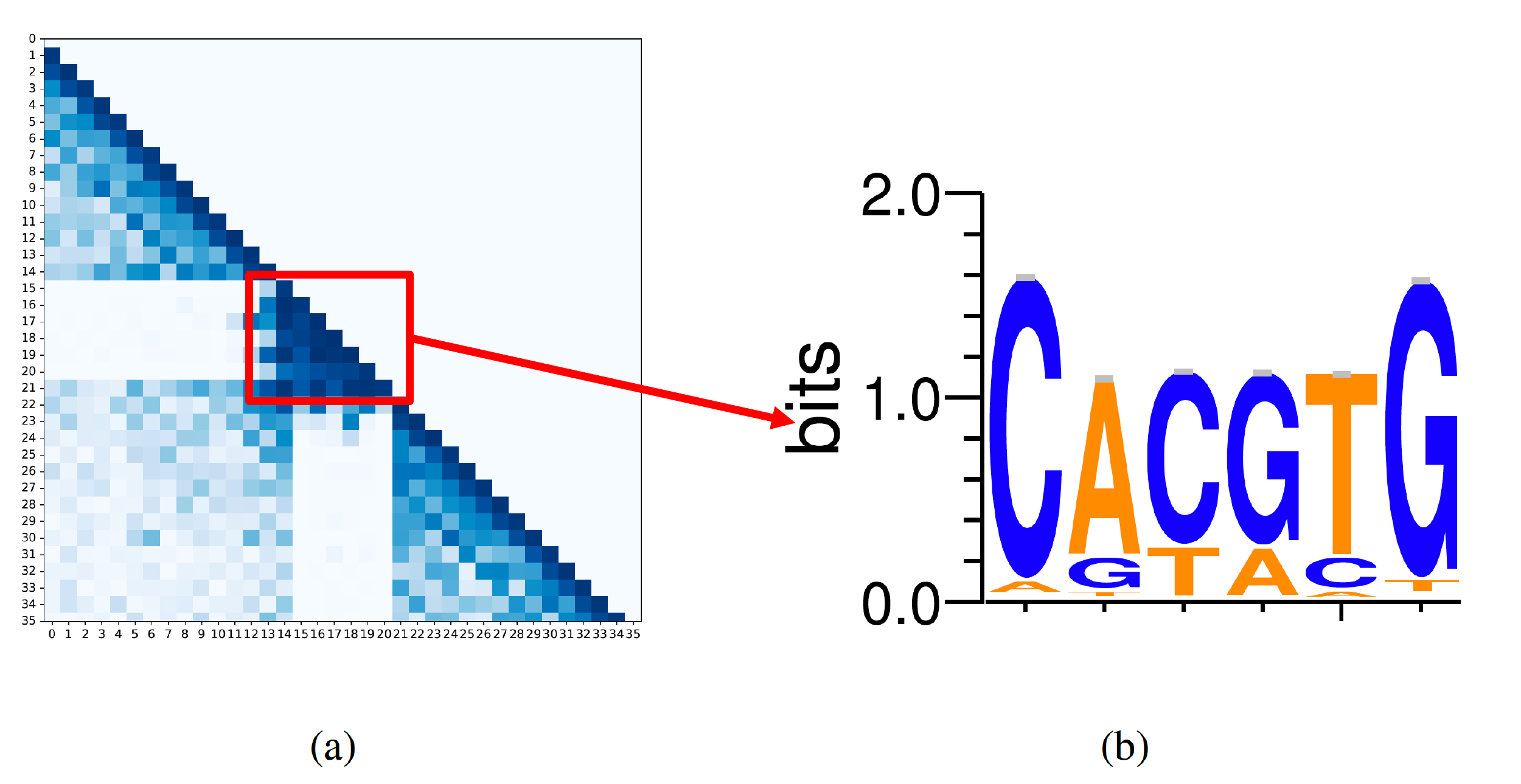}
\end{center}
\caption{
Dependency Graph for MYC bound DNA sequences learnt by M-\methodName on MYC dataset. We observe the motif ``CACGTG'' in the densely connected positions $16$ to $22$ which  forms a canonical DNA sequence\cite{staiger1989cacgtg}.
\label{fig:myc}
}
\end{figure*}

\subsection{DNA MYC Binding Data}
We use the MYC binding DNA data\footnote{\url{https://www.ncbi.nlm.nih.gov/geo/query/acc.cgi?acc=GSE47026}} about DNA binding specificity of a transcription factor MYC. 
We use the bound sequences as positive class and the unbound sequences as negative class. We split the data into 80/10/10. Figure~\ref{fig:myc} shows the learnt dependency structure for bound sequences. We observe a densely connected subgraph for positions $16$ to $22$. We show the corresponding sequence logo for the test data bound sequences. We observe the motif CACGTG which which forms a canonical DNA sequence\cite{staiger1989cacgtg}, also shown by \cite{tsang2020feature}. We also observe that in general the adjacent sequential positions depend on each other. 

{\bf Hyperparameters}: We use $struct_H=128$, $l_{sparse}=0.001$, $task_H=64$, $init_H=32$, $task_L=2$, $nheads=4$ and a learning rate of $0.001$. We use a batch size of $256$.

\subsection{Real World Dataset: Stanford Sentiment Treebank-2 Dataset}
We also evaluate \methodName on Stanford Sentiment Treebank (SST-2) binary sentiment classification task. We pad all sentences to a fixed length of 56.  Instead of one hot feature representation, we use pretrained GloVe\cite{pennington2014glove} embeddings both for the structure learner as well as the task learner. 

{\bf Hyperparameters}: For this dataset, we use a shared structure learner MLP with two layers with hidden size $\{1024,512\}$, $l_{sparse}=0.01$, $task_H=128$, $init_H=100$, $task_L=4$, $nheads=4$ structure learner dropout of 0.3, task learner dropout 0.2  and a learning rate of $0.0001$. We use a batch size of $128$. For the M-\methodName variation, we use $task_H=64$. For GCN and MLP variations, we use $task_H=1024$. 
 We show the learnt graph in Figure~\ref{fig:text_hm}. Here, the color indicates belief in the presence or absence of an edge. Red indicates
probability of an edge; Blue indicates belief in the absence of an edge and yellow indicates maximum
uncertainty. Local neighbors depend on each other(indicated by the thick orange diagonal). The sentences are variable length accounting for the diagonal Y shape in the learnt dependency graph.

\begin{figure*}[tb]
    \centering
\begin{minipage}{.5\textwidth}    \includegraphics[width=.81\textwidth]{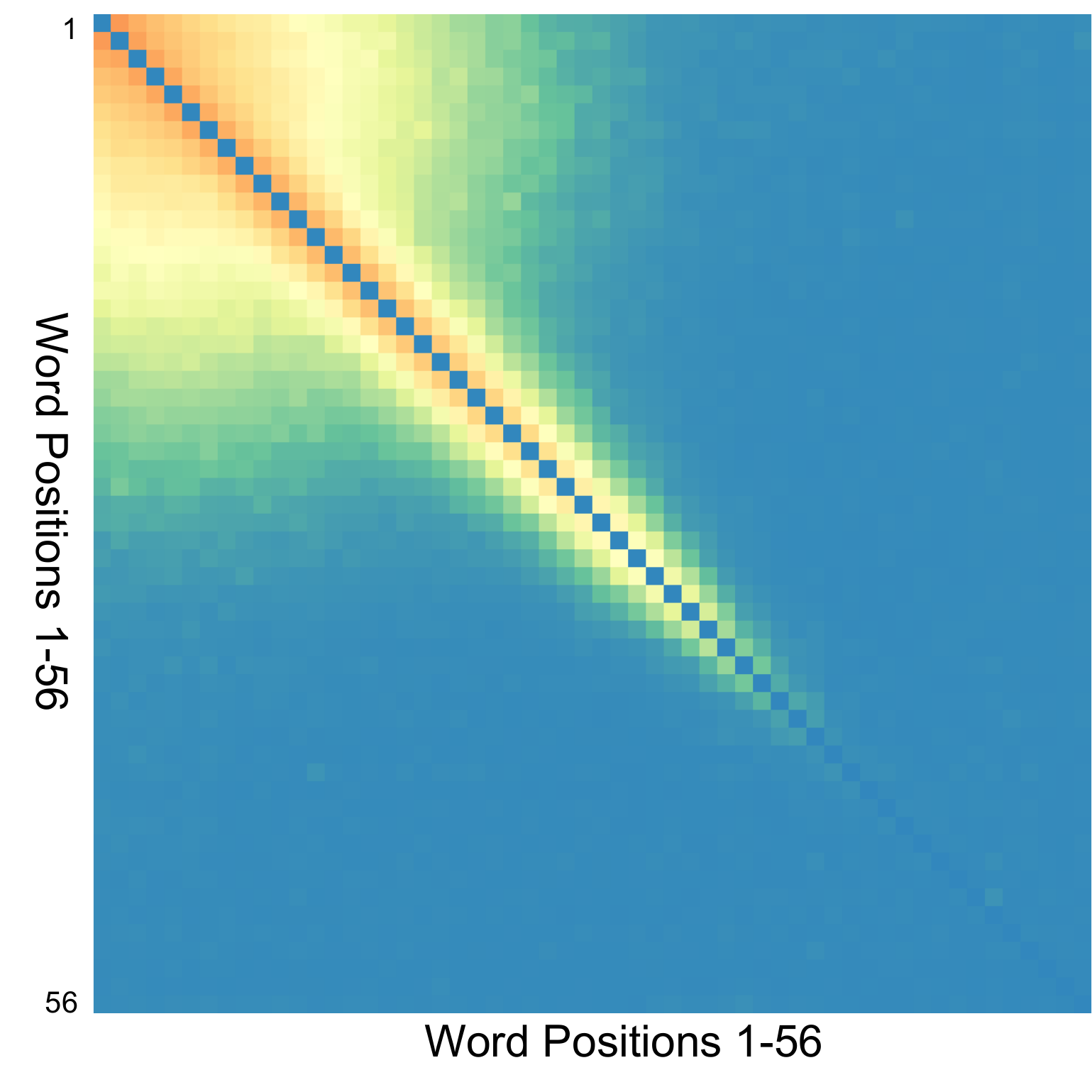}
\end{minipage}
\begin{minipage}{.39\textwidth}    
    \caption{ Graph learnt by \methodName on the sentiment classification SST-2 dataset.  Red indicates
belief in the presence of an edge; Blue indicates belief in the absence of an edge. In the learnt graph by \methodName, local neighbors depend on each other(indicated by the thick orange diagonal). The sentences are variable length accounting for the diagonal Y shape.    \label{fig:text_hm}}
\end{minipage}
\hfill
\end{figure*}
}

\end{document}